# The Temporal Graph of Bitcoin Transactions

**Vahid Jalili**
jalili.vahid@proton.me

## Abstract

Since its 2009 genesis block, the Bitcoin network has processed $>1.08$ billion (B) transactions representing $>8.72$B BTC, offering rich potential for machine learning (ML); yet, its pseudonymity and obscured flow of funds inherent in its UTxO-based design, have rendered this data largely inaccessible for ML research. Addressing this gap, we present an ML-compatible graph modeling the Bitcoin's economic topology by reconstructing the flow of funds. This temporal, heterogeneous graph encompasses complete transaction history up to block $863\,000$, consisting of $>2.4$B nodes and $>39.72$B edges. Additionally, we provide custom sampling methods yielding node and edge feature vectors of sampled communities, tools to load and analyze the Bitcoin graph data within specialized graph databases, and ready-to-use database snapshots. This comprehensive dataset and toolkit empower the ML community to tackle Bitcoin's intricate ecosystem at scale, driving progress in applications such as anomaly detection, address classification, market analysis, and large-scale graph ML benchmarking. Dataset and code available at github.com/b1aab/eba

## 1 Introduction

The advent of Bitcoin introduced a decentralized public ledger where cryptographic rules govern issuance, secure ownership, prevent double-spending, and enable value transfer. Since its 2009 inception, as of block $863\,000$, its ledger publicly records $>1.08$B transactions involving $>1.32$B unique addresses and representing $>8.72$B BTC in cumulative transferred value. Spanning $>16$ years of real-world economic activity, its ledger is a unique resource for machine learning research. It enables diverse applications (surveyed in [1]), including: network security analysis such as anomaly detection and fraud prevention; entity analysis like de-anonymization (identifying exchanges, miners, gambling sites) and linking pseudonymous addresses via behavioral patterns; economic modeling such as market prediction and value flow analysis; multimodal studies connecting on-chain activity to real-world events for socio-economic insights; and benchmarking for ML methods on large graph datasets.

Despite this rich potential, Bitcoin's design, particularly its Unspent Transaction Output (UTxO) model and pseudonymity (i.e., identified by a cryptographic identity, such as a public key, instead of a real-world identity or complete anonymization), poses significant challenges for ML applications. The UTxO model represents funds as values associated with atomic transaction outputs that only an intended recipient can spend. Consequently, unlike account-based systems, there is no readily accessible *state* (e.g., account balance, transaction count, or counterparties) for an entity (e.g., an individual, organization, or service). Reconstructing this entity's state requires tracing and aggregating numerous UTxOs across different blocks. Pseudonymity compounds this: transactions link to cryptographic addresses, and single entities often control numerous addresses. Mapping these addresses to entities, which is essential for building reliable entity-level features, requires sophisticated heuristics or external data [2–4]. Common privacy practices, like using new addresses per transaction or techniques like CoinJoin [5], deliberately further complicate efforts to resolve entities' state. Effectively applying ML to Bitcoin's ledger therefore demands specialized feature engineering.

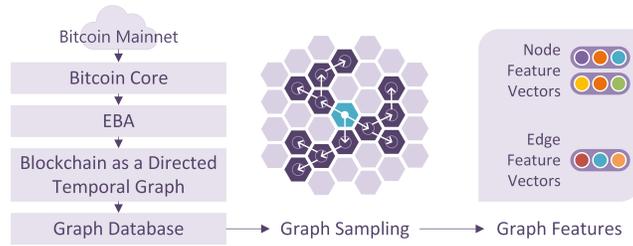

Figure 1: EBA interfaces with the Bitcoin *Mainnet* via Bitcoin Core to fetch and unpack block, transaction, and script information. EBA models the extracted data as a single directed temporal heterogeneous graph composed of multiple node and edge types, aiming for completeness and ease of use in machine learning applications. The graph is serialized into TSV files and, for efficient querying, imported into specialized graph databases. Additionally, EBA implements configurable sampling algorithms, including an adaptation of the Forest Fire model [6, 7] (section A.2.1), to sample communities, which are subsequently represented as node and edge feature vectors.

We model Bitcoin's transactions as a heterogeneous temporal graph representing the flow of funds and the topology of transactions (i.e., various inputs and outputs of a transaction and the set of transactions in a block). The graph consists of four node types interconnected by six types of directed edges timestamped by block height, which model coin issuance, fund transfers, and topology of transactions, hence capturing the state of entities. Cycles can emerge within this representation, reflecting closed loops of economic exchange. Section 2 details longitudinal blockchain statistics, providing insights to guide the effective interpretation and application of the graph model. Subsequent sections detail graph construction methodology (section 3), its characteristics (section 4), and methods for sampling communities and subgraphs (section A.2). An overview of the methodology is presented in Fig. 1. All analyses presented herein are based on the blockchain data up to block height $863\,000$.

## 2 Statistical profiling of Bitcoin's on-chain metrics

Bitcoin maintains chronological order through a linked sequence of *blocks*, each encapsulating transactions and secured by a cryptographic hash computed from both its contained transactions and the hash of the preceding block; thus, forming a *blockchain*. The Bitcoin protocol enforces rules ensuring that blocks are generated at regular intervals by adjusting the mining *difficulty* (Figs. 2 and A.7). Each block records a Median Time-Past ($t$), calculated as the median of the last 11 blocks [8]. The difference between the $t$ of consecutive blocks indicates that blocks are generated at regular intervals of every $\approx 10\,\text{min}$ ($\mu = 9.59 \pm 13.45$, Fig. A.8 for the distribution; throughout the manuscript, we adopt the notation $\mu = x \pm y$, where $x$ and $y$ denote the *mean* and *standard deviation*, respectively).

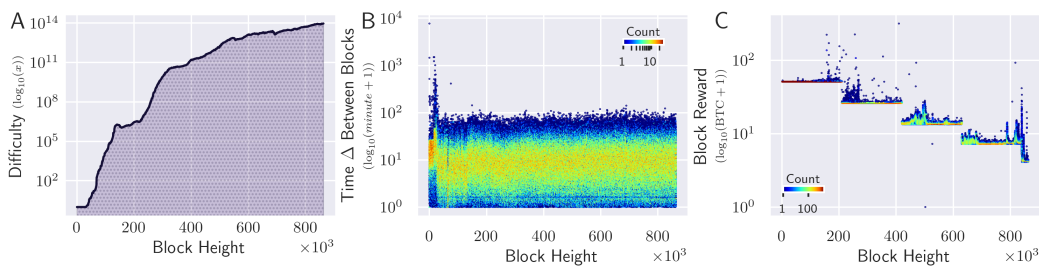

Figure 2: Chronology of Bitcoin's block generation dynamics: mining difficulty is adjusted (**A**, Fig. A.7) to regulate block generation to roughly every $10\,\text{min}$ (**B**, Fig. A.8), and the mining reward includes transaction fees and newly minted coins (**C**, Fig. A.6).

New coins are generated with each new block and awarded as incentives to the first *miner* (or mining pool) who successfully computes the cryptographic hash of the new block that meets the difficulty threshold (*mining*). Initially, 50 BTC were minted per block, halving every 210 000 blocks, totalling ≈19.76M BTC minted (Figs. 2, A.6 and A.19). The miner can claim both the minted coins and associated transaction fees; any rewards left unclaimed by miners become permanently unspendable. A total of 40.55 BTC in mining rewards went unclaimed across 116 067 blocks (entirely in block 501 726); including 21.82 minted BTC in 185 of these blocks (table A.5 and Fig. A.14).

Miners claim the newly generated coins by cryptographically locking them, typically using their own private-public key pairs, through a unique transaction present in every block called the *coinbase* transaction. To circulate these coins, miners must first unlock them, then re-lock them under new cryptographic conditions. Thus, transferring funds involves unlocking coins controlled by the sender and re-locking them such that only the intended recipient can unlock them. Consequently, each block encapsulates multiple transactions (*Tx*), each consisting of inputs (*TxIn*) that unlock previously secured coins (UTxO), and outputs (*TxOut*) that lock these coins under new cryptographic conditions. Accordingly, *coins* are values associated with atomic TxOut, each of which can be spent exactly once.

The >19.76 million minted coins were traded across >1.08B Txs, cumulatively amounting to >8.72B BTC exchanged. Coinbase transactions are the initial distribution point for these coins (Fig. A.21) and predominantly feature a single TxOut (93.2 %). While 1.6 % have ≥10 TxOut (typically associated with mining pools directly distributing rewards through coinbase Tx), this practice has a declining trend, reflected in the average number of coinbase TxOut dropping from $\mu = 4.07 \pm 23.87$ before block 500K to $\mu = 1.13 \pm 0.0$ afterward (Fig. 3). Beyond the coinbase Tx, each block contains $\mu = 1\,258.9 \pm 1\,285.6$ transfer Txs (Fig. 3). While the total number of TxIn ($\mu = 3\,161.21 \pm 2\,837.0$) and TxOut ($\mu = 3\,373.2 \pm 3\,448.9$) summed per block exhibit a positive correlation (Pearson $\rho = 0.77$), the average number of TxIn ($\mu = 2.9 \pm 18.2$) and TxOut per Tx ($\mu = 2.5 \pm 3.0$) across blocks show negligible linear correlation (Pearson $\rho = 0.02$, Fig. A.12). Despite a general increase in Txs per block with Bitcoin's adoption, "empty" blocks (that contain only the coinbase Tx and generated when a miner succeeds mining a block before including any Tx from the *transaction pool*) still occur, accounting for 10.4 % (89 714) of all blocks (Fig. 3).

Spending patterns for the >8.72B BTC total transferred value have evolved significantly (Fig. 3). For instance, the total BTC transferred per block increased dramatically from $\mu = 0.59 \pm 4.6$ in the first 300k blocks to $\mu = 5\,746.7 \pm 6\,054.8$ in the latest 300k blocks (Fig. A.13). Complementing the volume, the age of spent TxOut provides insights into holding behavior. Let $\delta_\mu$ be the average age of spent TxOut in a block, defined as the difference between the block heights at which a TxOut is spent and created, where $\delta_\mu = 0$ means the TxOut is spent in the same block in which it is created. Minted coins have specific spending constraints: they must mature for at least 100 blocks before being spent. On average, spent minted coins have an age of $\mu = 25k \pm 65k$ blocks, with each block spending $\mu = 64.8 \pm 305.7$ such coins (Fig. A.18). Of the >19.76M minted coins, >1.76M remain unspent; the >17.98M that have entered circulation contribute to the >8.72B BTC total exchange volume (Figs. 4 and A.19). Unlike newly minted coins, regular TxOut can be spent within the same block they are created. In 703 940 *non-empty* blocks, at least one TxOut was spent in the same block where it was created. Excluding *empty* blocks, the mean $\delta_\mu$ over the most recent 200k blocks is $\mu = 5\,298.0 \pm 10\,739.4$. Considering $\delta_\mu$ in two consecutive blocks (lag 1) reveals three distinct spending patterns (Fig. 4C): (a) In 8.02 % of blocks, $\delta_\mu = 0$ in two consecutive blocks; (b) In 2.3 % of blocks, $\delta_\mu = 0$ with $\delta_\mu > 0$ in a proceeding block; and, (c) In 40.6 % of blocks, $\delta_\mu$ across two consecutive blocks falls within the 1k to 10k range.

Bitcoin utilizes a Forth-like scripting system as its [un]locking mechanism to define spending conditions [9]; TxOut contain the locking script, and TxIn provide the unlocking script or data. The scripts, in most cases, are referred to by their user-friendly representations, known as *addresses*. ≈1.32B unique addresses have been created, with each block referencing $\mu = 3\,373.5 \pm 3\,445.5$ addresses, including $\mu = 1\,538.8 \pm 1\,618.3$ newly generated ones (Figs. 3, A.11 and A.15). The shifting popularity among the nine common script types—from early dominance by public keys (P2PK) towards Pay-to-Public-Key-Hash (P2PKH) and, more recently, increased adoption of Pay-to-Witness-Public-Key-Hash (P2WPKH) [10]—highlights evolving preferences regarding privacy, security, and transaction efficiency (Figs. 3 and A.10). Nevertheless, P2PKH remains predominant over the entire blockchain, accounting for 44.9 % of all TxIn and TxOut.

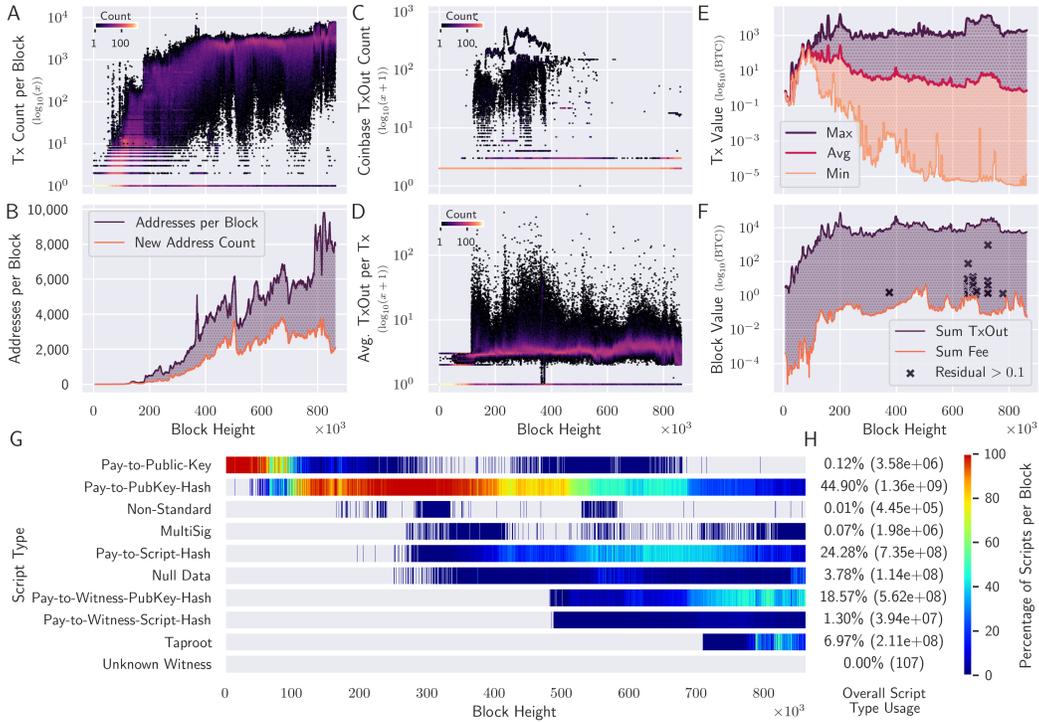

Figure 3: Longitudinal trends in Bitcoin Txs. Tx counts per block have increased (**A**) despite $10.4\,\%$ empty blocks, involving numerous new and reused addresses (**B**, Figs. A.11, A.12 and A.15). These addresses reference underlying scripts, and while their type usage has evolved (**G**, Fig. A.10), P2PKH remains the predominant type overall (**H**). Patterns of distributing minted coins have been evolving (**C**); similarly, while the transfer TxOut count per Tx evolves, it remains low across most Txs (**D**, Fig. A.12). Panel **E** shows per-Tx BTC volume trends: the max remains relatively stable, while average and minimum values decrease (more pronounced in the minimum, potentially reflecting exchange rates, Fig. A.13) Complementing panel **E**, **F** shows total BTC transferred and total mining rewards, revealing discrepancies where unclaimed TxIn value becomes lost funds (Fig. A.14). Rolling means in panels **B**, **E**, and **F** use a window of $5\,\text{K}$ blocks.

The byte size of each block is limited, initially $1\,\text{MB}$, and increased to $4\,\text{MB}$ with Segregated Witness (SegWit [10], Fig. A.9). Consequently, only a subset of transactions from the transaction pool (mempool) can be confirmed in each block, which limits overall network throughput and influences the inclusion of transactions with large or complex scripts.

While metrics like script type usage reflect evolving technical preferences, other on-chain metrics, such as TxOut spending patterns (*coin dormancy*), offer valuable insights into Bitcoin's economic behavior. For instance, low dormancy might suggest long-term holding, while high dormancy (or velocity) could indicate liquidation pressure. On-chain valuation metrics are invaluable barometers for various economic factors; although a deep dive into these economic indicators is beyond this paper's scope, we provide detailed statistics: Fig. 4 gives an overview of TxOut spending, with further details in Figs. A.16 to A.18.

## 3 Blockchain to graph

Blockchain technologies like Bitcoin implement a TxOut-centric model for on-chain data, which prioritizes immutability and double-spending prevention (fundamental requirements for distributed ledger technologies (DLT)) over explicitly modeling wallet-level activities. Consequently, performing wallet-level queries, such as identifying deposits/withdrawals within specific timeframes or analyzing longitudinal patterns of inter-wallet interaction, requires computationally intensive reconstruction from granular Tx/TxOut data, as wallet abstractions are not explicit on-chain. (While some Directed

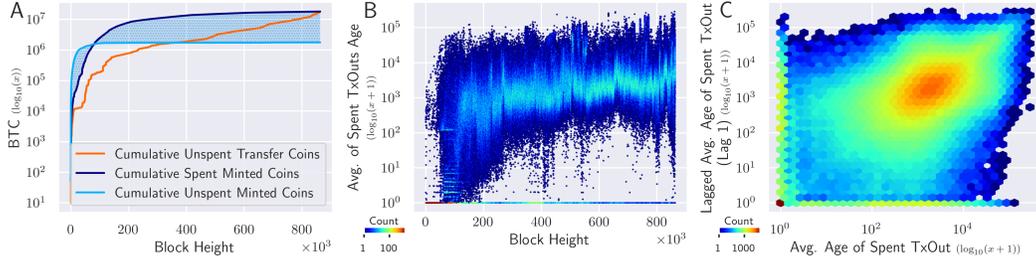

Figure 4: Bitcoin spending dynamics and coin age (dormancy). (**A**) Of >19.74 M minted coins, >1.76 M remain unspent, while the >17.98 M spent minted coins constitute the circulating funds (unspent non-coinbase TxOut, Fig. A.19), which are spent with varying holding periods: Panel (**B**) illustrates how the distribution of average spent TxOut age ($\delta_\mu$) varies across block height; Panel (**C**) compares consecutive $\delta_\mu$ values.

Acyclic Graph (DAG)-based DLTs offer on-chain wallet data, this research focuses on Bitcoin.) To facilitate richer analysis of fund circulation and economic behaviors within Bitcoin, we transform the TxOut-centric model into a wallet-centric graph model.

We construct this graph by including the transactional information and block metadata from the Bitcoin blockchain, while omitting purely cryptographic validation details. The graph is inherently heterogeneous, directed, temporal, and potentially cyclic. (See Fig. 5 for an overview; section A.1 details the encoding methodology, and node and edge encoding and their properties summarized in tables A.1 to A.4). The graph incorporates four node types: a single *Coinbase* node (minting origin), *Script* nodes (representing blockchain addresses, with incoming edges for received TxOut and outgoing edges for spent TxIn), *Tx* and *Block* nodes (providing context). Directed edges, stamped with block height, model blockchain relationships (e.g., value transfers, script-to-script) and structural links (e.g., confirmations, block-to-Tx, see Fig. 5). Critically, the *Tx* and *Block* nodes contextualize script interactions and serve as hypernodes, enabling the modeling of long-range dependencies or clustering among scripts, thereby supporting more sophisticated ML applications.

Unlike account-based systems, UTxO transactions consume a set of input UTxOs (controlled by potentially multiple scripts) and create a set of new output UTxOs (assigned to potentially multiple scripts), without explicitly linking individual sender to receiver scripts. To represent these flows more directly, our graph model introduces explicit script-to-script *Transfers* edges. Within each transaction, these edges form a complete bipartite graph connecting every input *Script* node to every output *Script* node involved (see table A.4 for details and examples in Figs. A.2 and A.3).

Our model distinguishes the two mining reward components: transaction fees and newly minted coins. While the Bitcoin protocol bundles both in the coinbase TxOut, our graph separates them to distinctly model value creation (minted coins) versus transfers of existing value (fees). Newly minted coins are modeled via *Mints* edges originating from the *Coinbase* node. In contrast, transaction fees are modeled using explicit script-to-script *Fee* edges connecting spender *Script* nodes to the miner's *Script* node (see table A.2 for details and examples in Figs. A.2 and A.3). These *Fee* edges exclusively model the on-chain transaction fees paid to miners and do not account for off-chain service fees, such as those an exchange might charge its users.

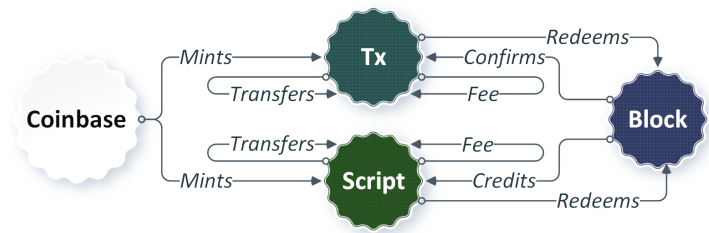

Figure 5: Schema of the graph. The process for creating the graph is detailed in section A.1, with nodes described in table A.1, and the mapping from blockchain properties to the graph provided in tables A.2 to A.4.

*Block* nodes provide temporal context, connecting to *Tx* nodes (*Redeems*, *Confirms* edges) and *Script* nodes (*Redeems*, *Credits* edges, Fig. 5). These directed edges enable efficient traversal of transaction histories along the block sequence: *Redeems* edges allow tracing funds in the spending direction, while *Credits* or *Confirms* edges allow tracing in the receiving direction. This structure serves two key purposes. First, it enables graph sampling methods to explore temporally co-occurring transactions (e.g., tracing funds received by a script and then sampling other transactions confirmed in the same block as that receiving event). Second, these *Block* nodes serve as crucial temporal anchors and structural elements for Graph Neural Networks (GNNs), enabling models to effectively learn time-dependent patterns and long-range dependencies or incorporate temporal encodings.

Overall, the graph is constructed with high fidelity to the Bitcoin blockchain. The extract, transform, load (ETL) pipeline parses blocks using Bitcoin Core (see Fig. 1), the protocol's reference implementation, which ensures that blocks are parsed accurately with respect to Bitcoin's long history of incremental protocol updates. The initial release of the graph is constructed from all transactions recorded in all blocks up to block $863\,000$, but it intentionally excludes: (a) cryptographic proofs (e.g., signatures), as their utility for ML tasks is unclear; (b) zero-value transactions; and (c) transactions with both $>20$ inputs and $>20$ outputs, as such transactions create extremely large bipartite graphs. Additionally, to preserve fidelity to the blockchain, the graph maps blockchain entities (blocks, transactions, scripts) to nodes and avoids aggregating distinct interactions; for instance, multiple transfers between the same entities within a block are preserved individually rather than summed. Self-transfers (change outputs) are also explicitly retained for completeness, even though they primarily serve a Bitcoin protocol requirement (see section A.1 for details).

The graph is provided in two formats: (a) nodes and edges in TSV files for broad compatibility, and (b) loaded into a specialized graph database for efficient querying. To facilitate common machine learning workflows, we provide tools for subgraph or community sampling using various algorithms, including an adaptation of the Forest Fire sampling method (section A.2.1). These tools also encode the sampled subgraphs into feature vectors (node and edge vectors) ready for input into ML models.

The graph, containing the complete history of Bitcoin transactions up to block $863\,000$, presents two challenges for users: the need for specialized computational resources due to its scale, and the requirement for regular updates given the volume of transactions added to the ledger on a daily basis. In its initial release, the compressed TSV files for the graph are $>1.17\,\text{TB}$ in size, expanding to $>3\,\text{TB}$ once imported into a specialized graph database. To make the dataset broadly accessible despite its scale, we have taken two steps. First, we provide several data access options for different use cases; for instance, we offer general-purpose sampled communities ($200\,\text{k}$ subgraphs, size $<1.5\,\text{GB}$) for exploratory applications that can be used without hosting a full graph database, and a database snapshot ($>700\,\text{GB}$) that facilitates starting a graph database instance without the computationally intensive import process. Second, to lower the memory requirements for large-scale graph analysis, the methods we provide are designed to run on systems with limited memory by relying on on-disk operations. Additionally, for regular updates, we provide methods to incrementally add transactions from new blocks after $863\,000$ to both the TSV files and the graph database, avoiding the need for frequent, full dataset rebuilds. Documentation for these resources is available on github.com/B1AAB/EBA.

## 4 Structural properties of the graph

The graph encodes Bitcoin's transaction history using $>2.4\text{B}$ nodes and $>39.72\text{B}$ block-height-annotated edges (Fig. 6). Over $1.3\text{B}$ *Script* nodes enable the script-level tracking of value transfer, with distinct edge types distinguishing between different categories of financial flows. Over $2.4\text{M}$ *Mints* edges connect *Script* nodes (belonging to miners or pools) to the single *Coinbase* node, modeling the initial reception of newly minted coins (Fig. A.21). Over $6.3\text{B}$ *Transfers* edges between the *Script* nodes model the circulation of existing funds. Among these, $20.3\,\%$ are self-transfers, representing the mechanism of returning the transaction remainder to the spender. Additionally, $>3.3\text{B}$ *Fee* edges model transaction fees paid to miners. These differ from *Transfers* edges as *Fee* edges represent payments to miners typically unrelated to the transaction parties, whereas *Transfers* are between potentially related sender/recipient entities.

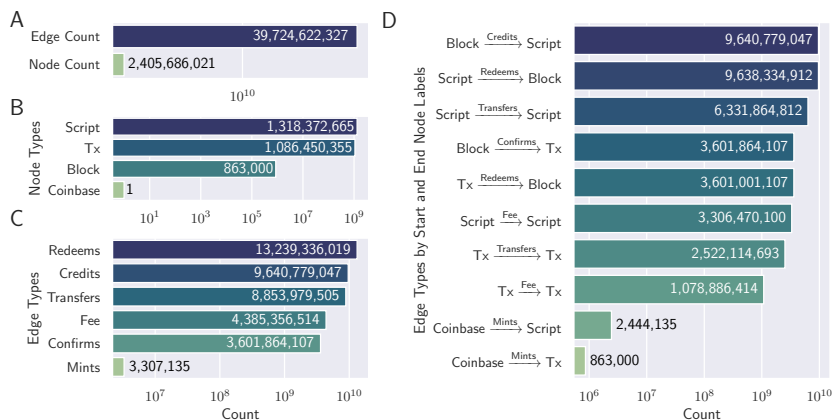

Figure 6: Statistical properties of the Bitcoin graph. (**A**) Total node and edge counts. Node (**B**) and (**C**) Edge type distribution. (**D**) Edge counts by relationship pattern: Source $\xrightarrow{\text{Edge Type}}$ Target.

Providing an aggregated perspective complementary to the script-level view, >1.08B *Tx* nodes represent individual transactions. Similar to the *Script* nodes, these *Tx* nodes are associated with distinct categories of financial flows, captured by three edge types: 863 000 *Mints* edges, >2.52B *Transfers* edges, and >1.07B *Fee* edges.

863 000 *Block* nodes anchor the graph's temporal dimension by connecting to *Tx* nodes via >3.6B *Redeems* and >3.6B *Confirms* edges, and to *Script* nodes via >9.6B *Redeems* and >9.6B *Credits* edges (Fig. 5). These *Block* nodes serve as crucial temporal reference points for modeling and sampling tasks (section 3).

Regarding degree distributions ($d_{\text{in}}$: indegree, $d_{\text{out}}$: outdegree), *Script* and *Tx* nodes exhibit similarities at lower degrees (Fig. 7). For *Script* nodes, 97.49 % have $d_{\text{in}} \leq 10$ and 92.56 % have $d_{\text{out}} \leq 10$, while 2.11 % have $10 < d_{\text{in}} \leq 100$ and 6.01 % have $10 < d_{\text{out}} \leq 100$. Similarly, for *Tx* nodes, 98.95 % have $d_{\text{in}} \leq 10$ and 98.06 % have $d_{\text{out}} \leq 10$, with only 0.83 % ($d_{\text{in}}$) and 1.69 % ($d_{\text{out}}$) falling between 10 and 100. However, the distributions diverge significantly in the tail (Fig. 7, Panel **D**). *Script* nodes show a pronounced long tail: 0.40 % have $d_{\text{in}} > 100$ (max $d_{\text{in}} \approx 282M$) and 1.43 % have $d_{\text{out}} > 100$ (max $d_{\text{out}} \approx 37M$). In contrast, 0.21 % ($d_{\text{in}}$) and 0.26 % ($d_{\text{out}}$) of *Tx* nodes exceed a degree of 100, with maximums reaching only $d_{\text{in}} \approx 8K$ and $d_{\text{out}} \approx 14K$. This difference is expected: *Script* nodes can be reused across multiple transactions throughout the blockchain's history, while each *Tx* node represents a unique transaction event confined to a single block. Despite this fundamental difference in lifespan and reusability, the total counts of *Script* nodes (1.31B) and *Tx* nodes (1.08B) remain relatively comparable. This is partly explained by the long degree tail for *Script* nodes, indicating that while many scripts adhere to privacy-preserving practices (infrequent reuse), a subset, often associated with exchanges or mining pools, are reused very frequently.

This temporal graph chronicles over a decade of Bitcoin's economic evolution. The discussed properties provide a foundation, but the detailed structure enables focused investigations into specific dynamics (e.g., longitudinal evolution of *n*-hop neighborhoods around degree-rich script nodes). While the dataset enables countless such targeted analyses, they are outside this paper's scope.

## 5 Applications and modeling considerations

The Bitcoin graph is a large-scale, real-world directed temporal graph modeling over 16 years of economic transactions. Its structure enables multi-scale analysis of on-chain economic activity, from modeling an entity's evolving trading patterns based on its multi-hop temporal neighborhood (node-level), to identifying communities of related (e.g., co-ownership) or coordinated activity (e.g., shared intent, subgraph-level). These characteristics make it a unique resource for applications in three key facets; first benchmarking graph algorithms and machine learning models. Computing metrics like eigenvector centrality, the longest path, or rich-club coefficient can provide significant insights into a real-world graph, particularly the Bitcoin ecosystem; however, performing these algorithms on a graph of this size is a significant challenge for most data science libraries, as they often rely on in-memory projections. This lack of scalability necessitates aggressive down-sampling or restriction to narrow temporal windows, which can compromise the relevance of the results. Similarly, this dataset can benchmark the scalability of graph ML pipelines; for instance, many neighborhood sampling implementations do not scale effectively, where sampling neighbors a few hops away for a given root node may require considerable computational resources. Beyond benchmarking for scalability, the generalization of ML methods across temporal windows can be tested against the Bitcoin graph's rapidly evolving nature, where network characteristics differ significantly across different temporal windows. Hence, the Bitcoin graph provides a benchmark for both the scalability of algorithms and ML models, and their robustness to temporal distribution shifts common in real-world networks.

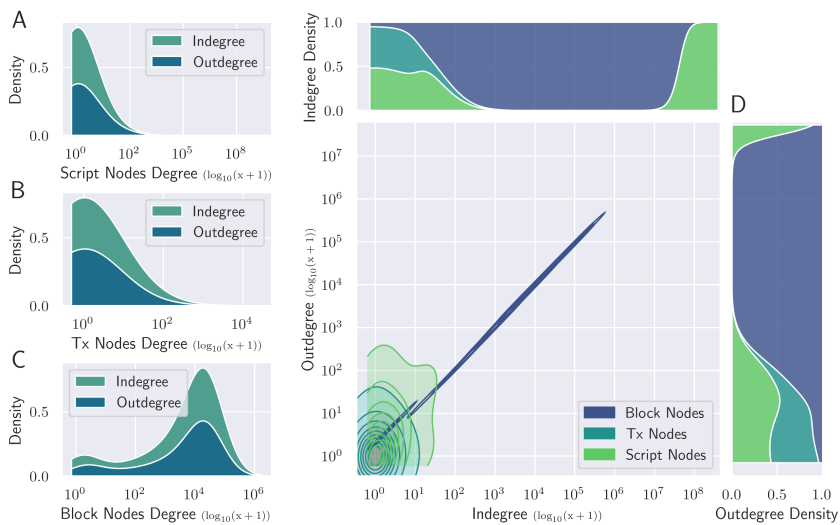

Figure 7: Bivariate degree distributions of the graph nodes, with degrees computed over all edge types and aggregated into bins of size 10 (lower bound). (**A**) The script nodes have low density ($5.57 \times 10^{-9}$) and variability (normalized Shannon entropy, $H_n = 0.24$, see table A.7), which aligns with many scripts being used a few times, alongside few high-degree scripts. (**B**) Tx nodes also have low density ($3.05 \times 10^{-9}$) and variability ($H_n = 0.21$), since they model unique transactions; their in-($\mu = 3.31 \pm 52.4$) and out-($\mu = 6.61 \pm 23.6$) degrees represent the input and output scripts per Tx. In contrast, Block nodes (**C**), fewer in number but acting as hypernodes, have higher density (0.018) and variability ($H_n = 0.87$). Panel **D** provides a comparison of these degree distributions.

Second, the Bitcoin graph bridges the gap between the cryptocurrency and ML communities. Its structure, scale, and temporal span make it ideal for training a base model for the Bitcoin ecosystem that learns on-chain trading patterns and the economic behavior of entities. In addition, the graph can be used for tasks such as node classification, edge prediction, community detection and classification, and subgraph matching. These tasks can introduce novel capabilities to cryptocurrencies, such as on-chain reputation systems, which could generate trust scores for wallets in applications like Decentralized Finance (DeFi) lending or, more proactively, enable fraud prevention through real-time risk assessments for wallets and transactions before they are submitted to the network. Finally, the graph can power AI agents for personalized cryptocurrency assistance, which could provide financial guidance by generating investment strategies that align a user's specified investment profile (e.g., risk tolerance) with on-chain dynamics.

Third, the Bitcoin graph can be extended to interdisciplinary applications when augmented with off-chain complementary data sources. For instance, a base model pre-trained on the Bitcoin graph can be fine-tuned for smart contract platforms to enable impactful DeFi applications, such as improving privacy-preserving credit reputation assessments to reduce risk in under-collateralized DeFi lending (i.e., users can borrow more funds than their deposited collateral). Additionally, the graph's temporal nature allows for synchronization with other temporal modalities; for example, combining on-chain activity with market indicators like Open-High-Low-Close (OHLC) can improve the predictive power of market forecasting models. Furthermore, analyses can correlate on-chain transactions with off-chain events (e.g., major policy announcements or sentiment shifts) to quantify their impact on the cryptocurrency ecosystem, which enables the study of the interplay between external events and economic behavior within decentralized systems.

Enabling these applications often requires modeling the true economic activity of entities as it is recorded on-chain through pseudonymous identities. Stemming from pseudonymity and the fact that a single entity can control many scripts, a foundational concept in cryptocurrency analysis is identifying script co-ownership using heuristics like address clustering, co-spending patterns, or change address identification. These heuristics are then used to model trading patterns for applications like anomaly detection. However, this analysis is confounded by privacy-enhancing techniques (e.g., CoinJoin) and heuristics of exchange services that may execute user trades from their own addresses to optimize on-chain activity. This "proxying" adds an additional layer of obfuscation; consequently, any application modeling an entity's trading patterns based on script co-ownership must also effectively account for these behaviors. Despite numerous research efforts [11, 2, 12] and proprietary solutions, entity resolution remains an active research area. The Bitcoin graph we present facilitates this research by enabling deep neighborhood analysis (many hops away) and augmentation with off-chain annotations, leveraging its detailed, temporal modeling of blocks, transactions, and scripts that closely mirrors the ledger's structure and preserves its temporal and economic context. We provide examples of building ML models on the Bitcoin graph, and augmenting the analysis with off-chain annotations at github.com/B1AAB/GraphStudio.

## 6 Limitations and future direction

While this research focuses on Bitcoin's transactional data, a natural extension is to include other DLTs and expand beyond transactional data (e.g., smart contracts) that expands the breadth of applications, including predictive or classification tasks on smart contracts. Additionally, the current graph is scoped to information explicitly available on-chain; including off-chain annotations such as entity type classification (e.g., exchange, miner, merchant), transaction intent (e.g., purchase, investment, mixing), or the distinction between illicit and non-illicit activity would significantly improve classification accuracy. Such annotations are provided by resources such as law enforcement agencies and cybersecurity reports; however, they generally cover a fraction of entities and are biased towards only actively investigated entities. Incorporating comprehensive off-chain annotations remains a key direction for future enhancements. Finally, the graph we presented here enables longitudinal study of the Bitcoin ecosystem (e.g., evolving or recurrent patterns, or periodically active communities, or negligible but consistent flow of funds inter- and intra-entities/communities); an in-depth analysis of the graph and application-specific ML solutions merit independent follow-up studies.

## 7 Ethical considerations

We build the graph exclusively using Bitcoin's on-chain data, which is publicly accessible and does not contain personally identifiable information or geographical locations, where entities are represented using cryptographic public keys with no direct association with real-world identities. We do not conduct de-anonymization or predictive modeling, or include off-chain annotations such as de-anonymized addresses. Since the graph encodes a ledger of high-volume real-world financial activities in its entirety, its analysis with sophisticated behavioral models combined with external annotations could be used to de-anonymize pseudonymous addresses and predict community behaviors. The outcome of such analysis can be used in fraud prevention or to identify illicit activities, or in other activities that could raise privacy concerns, result in discriminatory outcomes, or have broader socio-economic implications. We urge users to employ this dataset responsibly and adhere to the latest ethical best practices in all downstream studies.

## 8 Conclusion

We have introduced a graph encoding of Bitcoin's ledger, modeling the flow of funds from minting through all transactions, while omitting cryptographic validations. This temporal and heterogeneous graph comprises >2.4B nodes (four distinct types) and >39.72B edges (six distinct types), and preserves the chronological sequence of transactions by annotating all edges with block height. The graph models both the direct flow of funds between entities (represented by *Script* nodes), various input and output scripts in a transaction (*Tx* nodes), and the set of transactions in a block (*Block* nodes). Notably, *Block* nodes and *Tx* nodes function as contextual hubs or hypernodes that enable ML models to learn long-range dependencies, integrate temporal context, and identify co-occurring patterns. The defined edge types support diverse graph traversal strategies, such as tracing the spending of minted coin (e.g., Figs. A.5 and A.21), traversing neighborhoods of high-degree *Script* (often entities like exchanges), or analyzing the neighborhoods of *Script* nodes linked by common blocks ("temporal neighbors") or by common transactions (suggesting co-ownership or relatedness) to identify related activity.

Our in-depth statistical profiling of the Bitcoin blockchain provides context on its characteristics and evolving dynamics for effective interpretation and utilization of the graph. The graph is available as TSV files for broad compatibility and as a ready-to-use snapshot of a specialized graph database for enabling efficient querying and analytical tasks. Since a common practice for training ML models on a single large graph is training on various sampled subgraphs of the larger graph, we provide a suite of customizable sampling algorithms for sampling application-specific communities. Additionally, pre-sampled communities are provided to facilitate initial exploration and rapid prototyping.

Encoding the complete transaction history of the Bitcoin blockchain up to block $863\,000$ (spanning over 16 years of economic activity), this graph empowers the ML community to develop models for nuanced economic analysis, cryptocurrency-related applications, and provides a substantial real-world benchmark dataset for advancing large-scale graph machine learning research. Additionally, when integrated with other external temporal data sources (e.g., market data, or macroeconomic indicators), this graph enables interdisciplinary research, such as building models to explore and potentially predict broader socio-economic behaviors and trends influenced by or reflected in cryptocurrency activities.

# Supplementary Materials

## A.1 Blockchain to Graph

We model Bitcoin's blockchain as a single, directed temporal graph comprising nodes that represent scripts, transactions, and blocks, and block-height annotated (temporal) directed edges that aim for a complete representation of transactional data while also providing structural information that enhances machine learning tasks, such as modeling long-range dependencies and community structures, partly by leveraging block and transaction nodes as contextual hubs. This design intentionally replicates the blockchain data with high fidelity, which allows downstream analyses to either model the data with precision or, for simplicity, aggregate certain interactions (e.g., self-transfers or multiple transfers between the same scripts within a single transaction or block). Fig. A.1 provides a schematic of this modeling process, and the following sections outline the steps in detail.

Block 2 817 serves as an illustrative example; this block is notable as the first to include a transaction fee and also contains numerous self-transfers, making it an exemplary case for demonstrating our node and edge design decisions.

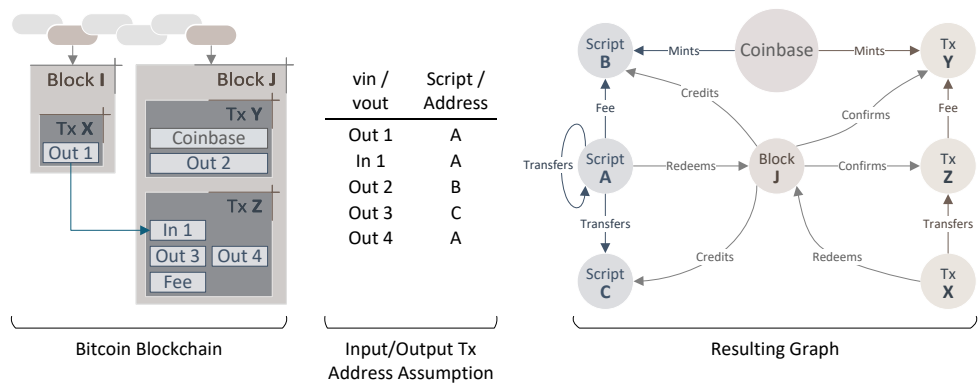

Figure A.1: Illustrative schematic of the graph model, constructed from Bitcoin's blocks, transactions, and scripts.

### A.1.1 Interfacing with the Bitcoin Network

To retrieve data, use the API endpoint of Bitcoin Core (widely recognized as the reference implementation). First, set the following in the Bitcoin Core configuration file, then start the node:

```
rpcbind=127.0.0.1
server=1
rest=1

# Creates an index of transactions in the blockchain.
txindex=1

# Increases the work queue depth so the client can
# respond to more concurrent API requests (default is 64).
rpcworkqueue=128
```

### A.1.2 Retrieving Block Data

Bitcoin Core retrieves block data given a block hash. Therefore, to obtain a block at a specific height, a two-step API process is necessary: first, an API request retrieves the block hash for the given height; second, this hash is used to fetch the actual block data. The API requests for this process are as follows:

1. Get block hash.

```
GET rest/blockhashbyheight/2817.hex HTTP/1.1
Host: localhost:8332
Content-Type: text/plain

00000000d50a3cd05e451166e5f618c76cc3273104608fe424835ae5c0d47db9
```

2. **Get block data.** In the following examples, API responses from the Bitcoin Core client are presented in `JSON` format for intuitive readability. While block data is generally processed and communicated as byte arrays by client applications within the network (often displayed in its `raw` hexadecimal-encoded form), interpreting these raw byte arrays directly requires significant domain knowledge to map specific byte sequences to their corresponding attributes due to its schema-less serialization. We therefore use the `JSON` representation, as its key-value pair structure is self-describing and avoids the need for such manual annotation of raw hexadecimal data. For brevity in the examples, some key-value pairs are replaced with `"*":"*"`, and long strings are truncated, with omitted parts marked by an asterisk (∗). Each block contains multiple Txs, with the count specified by the `nTx` key and the list of transactions under the `tx` key. Every transaction comprises three main attribute groups:

   (a) Metadata, such as `txid` or `hash`;
   (b) A list of inputs (TxIn, denoted `vin`); and
   (c) A list of outputs (TxOut, denoted `vout`).

```
GET /rest/block/00000000d5*db9.json HTTP/1.1
Host: localhost:8332
Content-Type: application/json

{
  "hash": "00000000d5*db9",
  "height": 2817,
  "*": "*",
  "nTx": 4,
  "tx": [
    {
      "*": "*",
      "in": [{"coinbase": "*", "sequence": 4294967295}],
      "vout": [
        {"value": 52.01, "n": 0, "scriptPubKey": {"*": "*"}}],
    },
    {
      "*": "*",
      "vin": [
      {
        "txid": "a87*",
        "vout": 1,
        "prevout": {"height": 2813, "value": 34.93, "*": "*"},
        "*": "*"
      }],
      "vout": [
        {"value": 1.0, "n": 0, "scriptPubKey": {"*", "*"}},
        {"value": 32.93, "n": 1, "scriptPubKey": {"*", "*"}}],
      "fee": 1.00000000,
      "hex": "0100000001db*"
    },
    "*" : "*"
  ]
}
```

### A.1.3 Deriving Addresses from Scripts

The inputs (TxIn) and outputs (TxOut) of a Bitcoin transaction (Tx) are defined by a Forth-like stack of `OP_CODEs`, known as Bitcoin *Script* (en.bitcoin.it/wiki/Script), which specifies the conditions required to spend funds. A Bitcoin Tx is considered *confirmed* if its script evaluates to *true* (i.e., upon execution, the combined locking script from an TxOut and unlocking script from an TxIn evaluates to *true*, or non-zero). There are several standard transaction script types, such as Pay-to-PubkeyHash (P2PKH), Pay-to-Script-Hash (P2SH), and Pay-to-Taproot (P2TR, Fig. A.10). An *address* is a user-friendly way of referencing a script, typically representing the hash of a specific part of standard scripts; however, not all scripts have an associated address.

Bitcoin Core extracts and returns addresses for most standard scripts (see the following example). For scripts where Bitcoin Core does not return an address, we use the NBitcoin library (github.com/MetacoSA/NBitcoin), which can derive an address for an extended set of scripts based on their type.

```
"scriptPubKey": {
  "asm": "OP_DUP OP_HASH160 6934efcef36903b5b45ebd1e5f862d1b63a99fa5 OP_EQUALVERIFY
    ↪  OP_CHECKSIG",
  "desc": "addr(1AbHNFdKJeVL8FRZyRZoiTzG9VCmzLrtvm)#64qpwnsw",
  "hex": "76a9146934efcef36903b5b45ebd1e5f862d1b63a99fa588ac",
  "address": "1AbHNFdKJeVL8FRZyRZoiTzG9VCmzLrtvm",
  "type": "pubkeyhash"
}
```

We reference a script using its address as a unique identifier if the address can be determined; otherwise, we reference it using a compound identifier composed of the following:

```
[Output Index in the Transaction]-[Transaction ID]
```

### A.1.4 Graph Components

We model the Bitcoin blockchain as a single directed temporal heterogeneous graph, where the block height is the temporal attribute. table A.1 summarizes the nodes in the graph, and the different types of edges between these nodes are discussed in the following sections.

### A.1.5 Coinbase Transaction to Graph

Each block begins with a unique coinbase transaction, created by the miner(s) of that block. This transaction is distinct as its input does not reference any previous TxOut; consequently, it is the only transaction that introduces new coins (minted coins) into circulation, as all other transactions merely transfer previously existing coins. The value of the coinbase Tx represents the total mining incentive for that block, comprising both the protocol-defined newly minted coins and the sum of all transaction fees paid by other transactions included in the block. While this total value represents the maximum the miner can claim (and distribute to any number of TxOut), miners have occasionally claimed less, resulting in the unclaimed portion becoming permanently lost (see table A.5 and Fig. A.14 for details). The TxOut scripts of the coinbase Tx are determined by the miner(s), typically directing funds to their own wallets.

To model the input of the coinbase Tx, we define a unique *Coinbase*—the only such node in the entire graph. This node has no incoming edges, and its outgoing edges connect exclusively to nodes representing the mining process (Fig. A.21).

In the Bitcoin blockchain, the output values of a coinbase Tx represent the sum of the mining reward (newly minted coins) and the total fees from other transactions within the block (circulating coins). However, our model treats minted and circulating coins separately. Accordingly, the graph includes distinct edges for minted coins and fees, as detailed in table A.2. Fig. A.2 provides an example of the graph modeling the coinbase Tx of the block at height $2\,817$.

### A.1.6 Non-Coinbase Transactions to Graph

We refer to every Tx in a block other than the first one (i.e., the coinbase Tx) as a non-coinbase transaction. Although non-coinbase transactions share several similarities with coinbase Tx, they have key differences:

```
1   "tx": [
2     {
3       "txid": "e95",
4       "vin": [{"coinbase": "..."}],
5       "vout": [{"value": 52.01}]
6     },
7     {
8       "txid": "f8b"
9       "vin": [{"txid": "a87", "vout": 1, "prevout": {"value": 34.93}}],
10      "fee": 1.0,
11    },
12    {
13      "txid": "65f"
14      "vin": [
15        {"txid": "f8b", "vout": 0, "prevout": {"value": 1.0},
16        {"txid": "f8b", "vout": 1, "prevout": {"value": 32.93}
17      ],
18      "fee": 1.0,
19    },
20    {
21      "txid": "5b6"
22      "vin": [
23        {"txid": "65f", "vout": 0, "prevout": {"value": 1.0},
24        {"txid": "65f", "vout": 1, "prevout": {"value": 31.93}
25      ],
26      "fee": 0.01,
27    }
28  ]
```

(a) A simplified JSON object representing the block at height 2 817, including minimal transactional information about fees and mining.

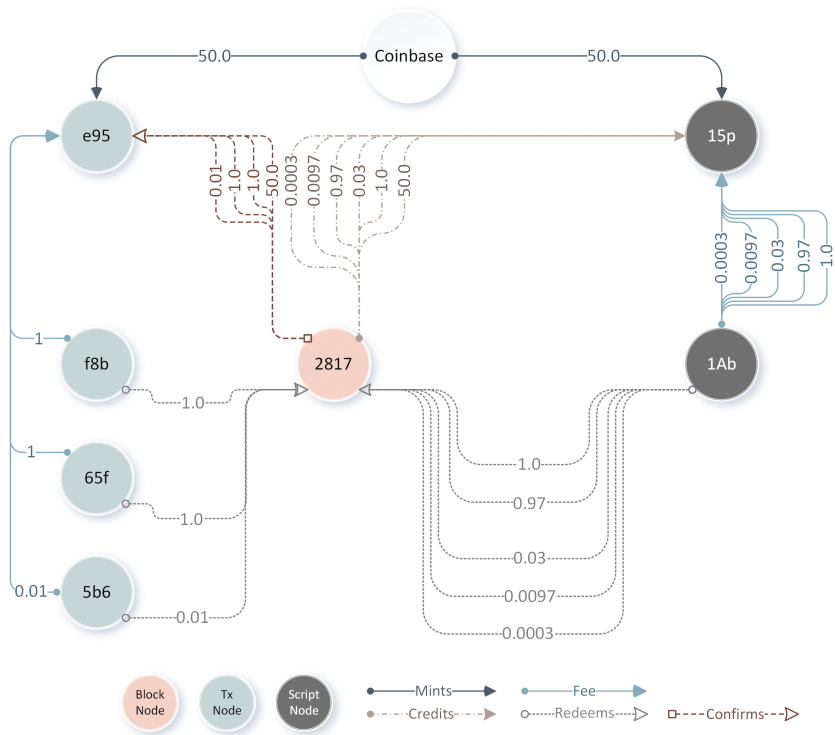

(b) The graph modeling the fee and mining Txs of the block at height 2 817.

Figure A.2: An example of modeling the coinbase Tx as a graph, which pays the mining reward to the miners, where the reward is the sum of the minted coins (generated) and fee (circulating coins).

Table A.1: The nodes of the graph.

| Node Label | Description | Properties |
|---|---|---|
| Coinbase | A unique node with a single instance in the entire graph used to model the coinbase Tx section A.1.5. This node has no incoming edges, and its outgoing edges connect to other nodes representing mining. | None |
| *Script* Node | Models a script that is uniquely identified with its address (see section A.1.3). TxIn and TxOut of Txs are scripts, hence a script node models the TxIn and TxOut of Txs. | Address (string), Script type (categorical) |
| Transaction (*Tx*) Node | "Transfers" in the blockchain are organized in Txs, highlighting the relatedness and shared activity of inputs and outputs in a Tx. Hence, a transaction node aims to facilitate learning of such patterns and relationships between scripts. Additionally, the *Tx* node, together with other nodes and edges in the graph, ensures that the graph fully encapsulates the transactional information of the blockchain. | TxId (string), Size (int), VSize (int), Weight (int), Version (string), LockTime (int) |
| *Block* Node | Every edge has the block height attribute as its temporal attribute; hence, the graph provides the chronological information of transactions, even without a dedicated block node. However, a block node serves the role of temporal *anchor* or *super node*, enabling the models to capture long-range temporal dependencies and dynamics. Depending on the models and applications, block nodes (and their connected edges) can be omitted without compromising the graph's completeness. | Height (int), MedianTime (int), Confirmations (int), Difficulty (float), Transactions Count (int), Size (int), Stripped Size (int), Weight (int) |

- Each block contains exactly one coinbase Tx, whereas it can include an arbitrary number of non-coinbase Txs.
- The coinbase Tx generates new BTC, while non-coinbase Txs only circulate existing BTC.
- The input in the coinbase Tx is coinbase, a special input that does not reference any prior output, whereas the inputs of a non-coinbase Tx reference TxOut from other Txs.
- A coinbase Tx has exactly one input, while a non-coinbase Tx can have one or more inputs.

We model non-coinbase Txs similarly to coinbase Txs, with adjustments to account for these differences. In general, the process involves creating a node for each Tx, with an edge connecting it to the Tx that produced each referenced TxIn. Additionally, for each Tx, we create a complete bipartite graph between the TxIn and TxOut scripts, where the value property of each edge represents the proportion of an input's value that is "transferred" to the corresponding output. Table A.4 provides the list of edges used to model a non-coinbase Tx, and Fig. A.3 provides an example of modeling non-coinbase Txs, extending the example given in Fig. A.2 for modeling a coinbase Tx.

### A.1.7 Modeling scripts by their addresses yields greater insight than unique Tx in/out IDs

We model transaction inputs and outputs using the addresses derived from the scripts (see section A.1.3). This approach uses a single node to represent input and output scripts with the same address; its key advantage is providing a more informative modeling of the flow of funds.

An alternative approach would be to represent each input and output script separately using a unique identifier, such as `[Output index]-[Tx ID]`. Although this method more closely mirrors the transactions, it is less informative since transfers associated with a single script (e.g., a wallet) would be modeled as incoming and outgoing edges to separate nodes, making it difficult to infer

Table A.2: Nodes and edges modeling the transfer of mining rewards and fees to the miner(s). All edges share a common set of properties, including value (in BTC) and block height. Additionally, each edge has a corresponding edge connecting to a *Block* node, as detailed in table A.3.

| Edge Type | Description |
|---|---|
| Coinbase $\xrightarrow{\text{Mints}}$ Script | This edge models only the *minted* or *generated* coins (the mining reward is the total sum of minted coins and the collected fees). An edge of this type is created between the unique *Coinbase* node and every script node (see table A.1) representing each output in the *coinbase transaction*. The value of this edge is proportional to the output script's share of the mining reward. $$\text{Minted coins} = \text{Mining reward} - \text{Total fee}$$ $$\text{Value} = \text{Minted coins} * \frac{\text{BTC paid to the script}}{\text{Mining reward}}$$ |
| Coinbase $\xrightarrow{\text{Mints}}$ Tx | This edge models the *minted* or *generated* coins only and is created between the unique *Coinbase* node and the transaction node (see table A.1) representing the *coinbase transaction*. Hence, there is only one such edge per block. The value of this edge equals the amount of minted coins in the block. |
| Script$_u$ $\xrightarrow{\text{Fee}}$ Script$_v$ | An edge of type *fee* that models the fee paid in a transaction. The source node represents an input script in the fee-paying transaction, and the target node represents an output script in the *mining transaction*. A **complete bipartite graph** is created between the input scripts of a fee-paying transaction and output scripts of the *mining transaction*. The value of this edge is determined as follows. $$u_{\text{fee share}} = \text{Tx fee} * \left( \frac{u_{\text{value}}}{\max(\text{Total Tx input}, 1)} \right)$$ $$e = \text{round}\left( u_{\text{fee share}} * \frac{v_{\text{value}}}{\text{Total paid to miner}} \right)$$ where $e$ is the amount of BTC of the edge connecting the source node $u$ and target node $v$. The rounding method is "round half away from zero" to match the method used in Bitcoin Core. |
| Tx$_u$ $\xrightarrow{\text{Fee}}$ Tx$_v$ | This edge models the fee-paying transaction by connecting the node representing the fee-paying transaction (Tx$_u$, see table A.1) to the node representing the *mining transaction* (Tx$_v$). The value of this edge equals the total reward paid in the transaction. |

```
1  "tx": [
2    {
3      "txid": "e95",
4      "vin": [{"coinbase": "..."}],
5      "vout": [{"value": 52.01}]
6    },
7    {
8      "txid": "f8b"
9      "vin": [{"txid": "a87", "vout": 1, "prevout": {"value": 34.93}}],
10     "vout": [{"n": 0, "value": 1.0}, {"n": 1, "value": 32.93}]
11     "fee": 1.0,
12   },
13   {
14     "txid": "65f"
15     "vin": [
16       {"txid": "f8b", "vout": 0, "prevout": {"value": 1.0}},
17       {"txid": "f8b", "vout": 1, "prevout": {"value": 32.93}}
18     ],
19     "vout": [{"n": 0, "value": 1.0}, {"n": 1, "value": 31.93}]
20     "fee": 1.0,
21   },
22   {
23     "txid": "5b6"
24     "vin": [
25       {"txid": "65f", "vout": 0, "prevout": {"value": 1.0}},
26       {"txid": "65f", "vout": 1, "prevout": {"value": 31.93}}
27     ],
28     "vout": [{"n": 0, "value": 0.01}, {"n": 1, "value": 32.91}]
29     "fee": 0.01,
30   }
31 ]
```

(a) A simplified JSON object representing the block at height 2 817, including minimal transactional information about the flow of circulating funds.

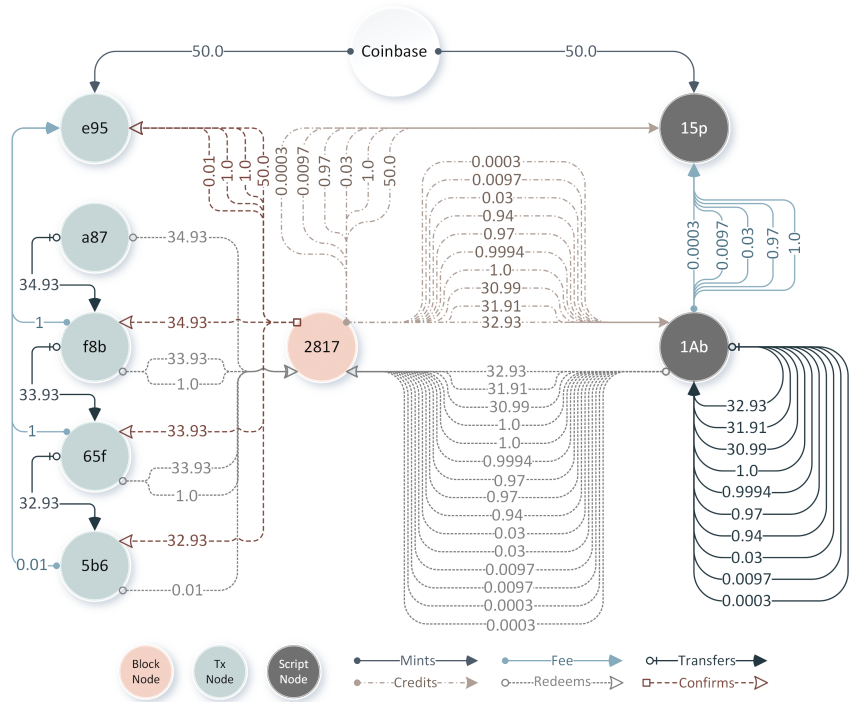

(b) The graph modeling all the transactions of the block at height 2817.

Figure A.3: An example of modeling the flow of generated and circulating funds in a block (height 2 817). This example extends on the example provided in Fig. A.2, which included only the flow of funds collected by the miner by incorporating the edges that model the *transfer* of funds between non-miner scripts.

Table A.3: For every incoming or outgoing edge of a *Script* or *Tx* node, a corresponding edge is defined between that node and the *Block* node. For instance, for every incoming edge of type *Fee* to a node $v$, an edge of type *Credits* is created between the *Block* node and $v$. Similarly, an edge of type *Redeems* is created for every outgoing edge of type *Fee* or *Transfers* from a *Script* or *Tx* node. The value of a *Redeems* edge equals the value of its corresponding *Fee* or *Transfers* edge. All edges share a common set of properties, including value (in BTC) and block height,

| Edge Type | Orientation | Corresponding Edge |
|---|---|---|
| $Script_u \xrightarrow{Fee} Script_v$ <br> $Script_u \xrightarrow{Transfers} Script_v$ | Outgoing | $Script_u \xrightarrow{Redeems} Block$ |
| $Tx_u \xrightarrow{Fee} Tx_v$ <br> $Tx_u \xrightarrow{Transfers} Tx_v$ | Outgoing | $Tx_u \xrightarrow{Redeems} Block$ |
| $Tx_u \xrightarrow{Fee} Tx_v$ <br> $Tx_u \xrightarrow{Transfers} Tx_v$ | Incoming | $Block \xrightarrow{Confirms} Tx_v$ |
| $Script_u \xrightarrow{Fee} Script_v$ <br> $Script_u \xrightarrow{Transfers} Script_v$ | Incoming | $Block \xrightarrow{Credits} Script_v$ |

that the funds are transferred in and out of the same wallet. Fig. A.4 illustrates these differences. The advantage of our script modeling approach is particularly pronounced in graph-based machine learning applications, as it enables capture patterns, communities, and flow of funds more accurately.

### A.1.8 Persisting and Querying the Graph

We model and serialize each block as an independent graph. We persist the nodes and edges of these graphs in character-delimited files, grouping them by node and edge types and storing them in independent batches of user-defined size. This organization makes it easier to study and extract subsets of the graph; for example, studying the nodes and edges representing transactions from blocks with heights ranging from $400\,000$ to $600\,000$.

While storing the graph in plain text files simplifies using it and supports a wide range of applications; however, running in-depth analyses can be challenging or even infeasible using plain text. For instance, finding all neighbor addresses at a distance of $h$ hops from a given address, or finding nodes with a given in-degree and retrieving their communities within $h$ hops, are both difficult tasks to perform efficiently on plain text files. Graph-based modeling enables the study of highly insightful aspects of blockchain data; however, the limited capacity of plain text serialization to support such analyses efficiently, hinders us from capturing the most compelling patterns, communities, and flows of funds.

To address these limitations, we leverage specialized graph database solutions designed for such queries. Numerous graph database options are available, ranging from managed cloud-based services like Amazon Neptune to self-hosted alternatives. We use Neo4j (neo4j.com) since it offers a self-hosted, free community edition. We provide (1) Solutions that format the nodes and edges into a format compatible with Neo4j; (2) Solutions for importing data, both in bulk and transactionally, into a Neo4j database; (3) Cypher queries (a graph-specific query language) to derive statistical insights from the graph; and (4) An automated method for sampling communities from the graph.

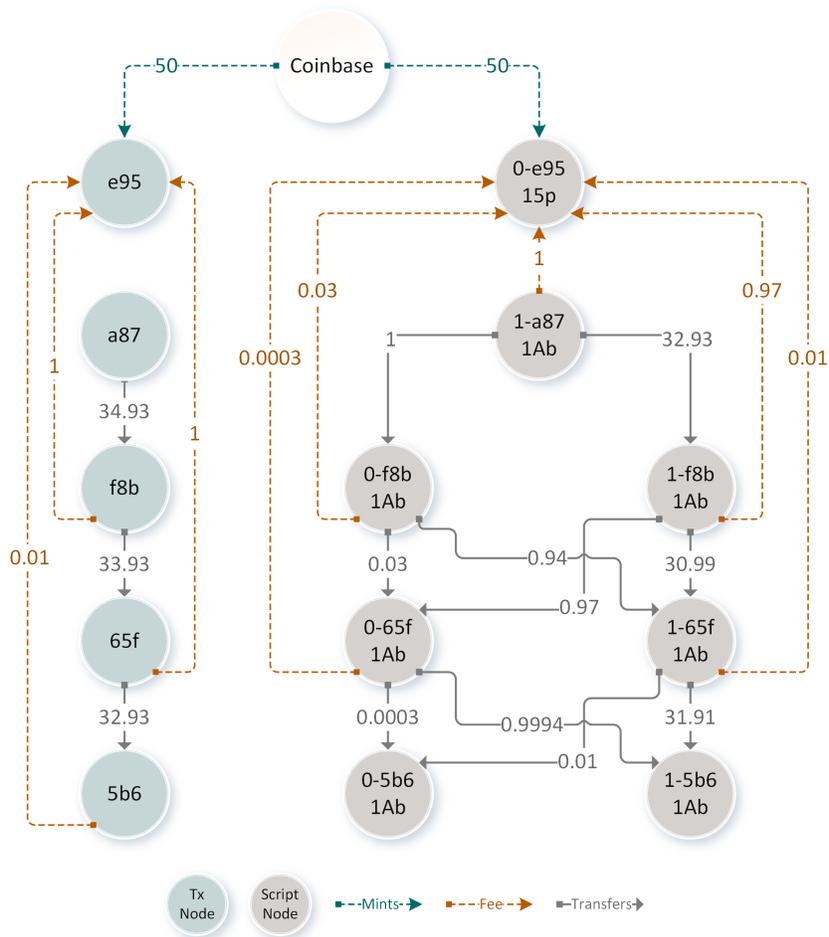

Figure A.4: This figure illustrates the same transaction as that modeled in panel (b) of Fig. A.3 (the block node and its related edges are omitted here to simplify the illustration). The graph in this figure uses unique IDs for transaction input and output scripts (see section A.1.7), whereas the graph in panel (b) of Fig. A.3 uses addresses derived from those scripts. The benefit of our modeling choice (i.e., the latter) becomes evident when comparing the two figures: the latter shows that funds are transferred into and out of a single script, while the former shows that funds are transferred between seven separate nodes/scripts. Our script modeling choice is particularly advantageous in graph-based machine learning applications, as it enables more accurate capturing and modeling patterns, communities, and, particularly, the flow of funds.

Table A.4: Nodes and edges modeling the circulation of funds. All edges share a common set of properties, including value (in BTC) and block height, and each edge is associated with a corresponding edge connecting to a *block* node, as given in table A.3.

| Edge Type | Description |
|---|---|
| $\text{Script}_u \xrightarrow{\text{Transfers}} \text{Script}_v$ | This edge models the transfer of funds between two *Script* nodes (see table A.1). The source node, $u$, represents the address of an input script within a transaction, and the target node, $v$, corresponds to the address of an output script in the same transaction. Since a Tx can contain multiple TxIn and TxOut, it is modeled as a complete bipartite graph linking each input node to every output node. A Tx is modeled with a complete bipartite graph of edges of type *Transfers* between its input script nodes and its output script nodes. Each edge's value is proportional to the fraction of the transaction's total input (excluding fees) provided by $u$ and the share of the total input allocated to $v$. In other words, the edge's value represents the portion of funds transferred from $u$ to $v$ relative to the total funds exchanged in the transaction. Specifically, the edge value is computed as follows. $$\text{value} = \text{Output } v \text{ value } * \frac{\text{Input } u \text{ value}}{\Sigma(\text{Tx inputs }) - \text{ fee}}$$ |
| $\text{Tx}_u \xrightarrow{\text{Transfers}} \text{Tx}_v$ | This edge models the transfer of funds between two *Tx* nodes (see table A.1). The source node, $u$, represents the transaction from which a TxOut, $n$, is used as a TxIn in the current Tx, while the target node, $v$, represents the transaction where the TxOut $n$ is redeemed. The value of this edge is equal to the value of the TxOut $n$. Accordingly, we model a Tx in a block with a node representing the transaction itself and nodes representing the transactions referenced in its list of inputs, along with edges from the input transactions to the transaction itself, such that the sum of the values of all the edges equals the total value of the inputs referenced in the transaction. |

## A.2 Graph Sampling for Model Training

The Bitcoin graph is a single, large-scale graph that is both heterogeneous (in node and edge types) and temporal (spanning over a decade). Considering its scale, training machine learning models directly on the full graph is often infeasible due to computational constraints. Considering its heterogeneity and temporal span, while enabling a wide range of insightful analysis and machine learning applications, also necessitate careful data selection aligned with specific training goals, which is crucial for isolating relevant nodes/edges and mitigating inherent confounders, such as those arising from temporal shifts in Bitcoin transaction patterns. A common approach for training on such graphs involves using sampled subgraphs generated specifically for the training goal.

Selecting an optimal graph sampling strategy is highly application-dependent, guided by the learning task (e.g., link prediction, node classification, graph property prediction) and learning paradigm (supervised, unsupervised, or self-supervised). Beyond the overall strategy, sampling parameters—such as neighborhood size or traversal depth (hops) from root nodes—are equally application-specific. The goal of an ideal sampling method is to generate subgraphs that retain information relevant to the training objective without introducing overly complex or potentially confounding topologies.

These considerations are particularly critical for the Bitcoin graph, given its heterogeneity and significant temporal evolution over more than a decade, during which network characteristics, transaction patterns and volumes have changed. Failure to account for such temporal drift during sampling can introduce significant confounders. For example, consider an edge prediction task where the intended goal is to predict edges based on local network topology. If positive examples are inadvertently sampled primarily from early periods (e.g., with typically higher BTC traded per transaction) while negative examples are drawn mainly from more recent periods (e.g., with lower BTC traded per transaction), a model might primarily learn to classify edges using the confounding temporal characteristic (like transaction value) instead of the intended topological patterns.

To facilitate downstream applications and benchmarking using the Bitcoin graph, we provide configurable implementations of fundamental graph traversal and sampling algorithms:

- Breadth-First Search (BFS);
- Depth-First Search (DFS); and
- an adaptation of the Forest Fire algorithm (see section A.2.1).

These methods are parameterized, allowing users to generate datasets tailored to specific modeling requirements. The parameters include:

- Traversal depth (hops);
- Whitelisted or blacklisted node and edge types;
- Termination conditions (e.g., encountering specific node/edge types); and
- constraints on subgraph size (minimum/maximum node and edge counts).

For demonstration purposes, we provide sampled subgraphs suitable for an unsupervised subgraph property prediction task. Specifically, this task involves differentiating between fully connected subgraphs (where a path exists between any two nodes) and those comprising potentially disconnected components (i.e., a "forest"). The sampling script accordingly generates two sets of subgraphs assigned binary labels: *Connected Graph* or *Forest*.

For each sampled subgraph, nodes and edges are serialized as feature vectors, yielding three files per subgraph:

1. Node feature vectors;
2. Edge feature vectors;
3. A graph-level label: *Connected Graph* or *Forest*.

Additionally, a separate labels file provides a mapping between each generated subgraph and its corresponding label within the overall dataset.

### A.2.1 Sampling Algorithms

Consider a node with an out-degree of 100, where each of its neighbors also has a similar out-degree. If the goal is to sample 10 nodes from the neighborhood within 3 hops, a BFS traversal will likely return 10 direct neighbors of the root node, and a DFS search will likely return the first neighbor found and 9 of that neighbor's neighbors. Therefore, neither of these methods will effectively sample nodes up to 3 hops away, nor will the sampled subgraph be representative of the broader 3-hop neighborhood structure.

Therefore, to sample representative neighborhoods in the Bitcoin graph, we developed a method that randomly selects a subset of neighbors for a root node. Then, for each of those selected nodes, it chooses a subset of their immediate neighbors, and continues this process until a termination criterion is met. The method is detailed in algorithm 1.

This approach aims to sample representative neighborhood subgraphs from the Bitcoin graph, addressing the limitations of BFS/DFS mentioned previously. The method follows the general principles of the Forest Fire model and is an adaptation of the Forest Fire sampling algorithm [7]. We provide an implementation of this algorithm and example graphs sampled using this method. Fig. A.5 shows the neighbors of the *Coinbase* node sampled using this method.

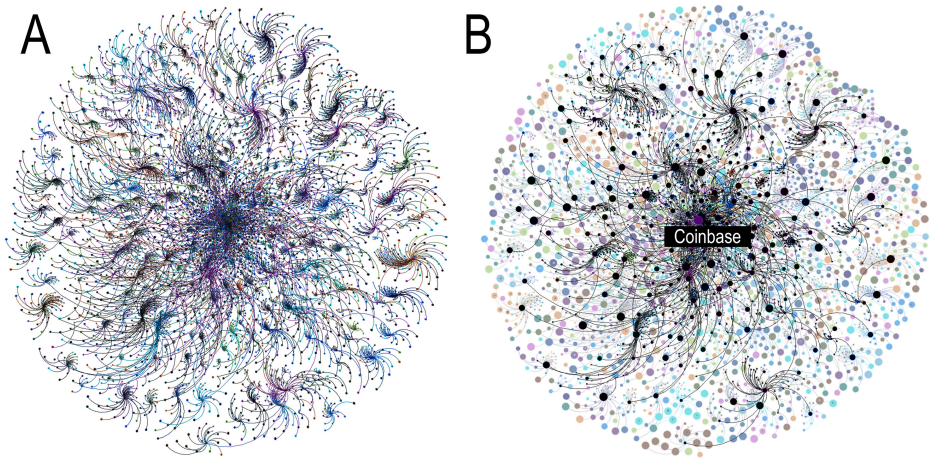

Figure A.5: Randomly sampled neighbors of the *Coinbase* node using algorithm 1 visualized using Graphistry (www.graphistry.com). (**A**) Shows all sampled nodes and edges. (**B**) Highlights outgoing edges originating from the *Coinbase* node within the 3-hop neighborhood.

## A.3 Monetary Units

Throughout this paper, the standard unit *BTC* is used to denote monetary units associated with transactions, which are modeled as edge values. With the increasing adoption of Bitcoin, there is a growing need to use fractions of a single coin for microtransactions. Currently, the smallest amount the protocol supports is $0.00000001$ BTC.

However, rounding errors in floating-point arithmetic can be challenging. To maintain consensus across the network among implementations in various programming languages, the Bitcoin Core application (the reference implementation of the protocol) represents monetary units as 64-bit integers. These integer units, commonly known as *satoshis*, are calculated by multiplying the BTC amount by $100\,000\,000$. Thus, $1$ BTC $= 100\,000\,000$ satoshis. Bitcoin Core converts this internal Satoshi representation back to BTC for reporting. To remain consistent with this practice, our method also uses Satoshi for internal calculations.

**Algorithm 1** Neighbor sampling algorithm, which randomly selects neighbors using Breadth-First Search (BFS) and iteratively expands the graph by applying a growth/decay factor to the number of neighbors sampled, until the maximum hop distance from the root node is reached.

1: **Input**
2:    $v_{\text{root}}$   The root node for neighbor sampling.
3:    $h_{\max}$   The maximum hops from $v_{\text{root}}$ to sample, aka. receptive field depth.
4:    $n$     The number of neighbors sampled from $v_{\text{root}}$ at the first hop.
5:    $\delta$     The decay/growth factor that adjusts the number of neighbors sampled at each hop, such that at hop $h$, a maximum of $n - h \times \delta$ neighbors are sampled.

6: **Output**
7:    $G$     Sampled graph

8: $V \leftarrow \emptyset$ , $E \leftarrow \emptyset$
9: TRAVERSEHOP($v_{\text{root}}, 0$)
10: **return** $G \coloneqq (V, E)$

11: **procedure** TRAVERSEHOP($v, h$)
12:    $G_v \leftarrow$ Get neighbors of $v$ at 0 hop with BFS algorithm. ▷ *Hop is fixed at 0 for simplicity, it can be dynamic and adjusted using a parameter similar to $\delta$.*
13:    $V'_v \leftarrow$ PROCESSSAMPLINGRESULTS($G_v, h$)
14:    **if** $h < h_{\max}$ **then**
15:       **for** each $v'$ in $V'_v$ **do**
16:          TRAVERSEHOP($v', h+1$)
17:       **end for**
18:    **end if**
19: **end procedure**

20: **procedure** PROCESSSAMPLINGRESULTS($G_v, h$)
21:    $V' \leftarrow \{\text{node} \in G_v\} \setminus V$
22:    $E' \leftarrow \{\text{edge} \in G_v\} \setminus E$
23:    $V' \leftarrow \text{Sample}(V', n - (h \times \delta))$     ▷ *Sample(S, n) randomly selects n items from set S.*
24:    $V \leftarrow V \cup V'$
25:    $V_{\text{new}} \leftarrow \emptyset$
26:    **for** each $e \in E'$ **do**
27:       **if** (target node of $e$) $\in V'$ **then**
28:          $E \leftarrow E \cup e$
29:          $V_{\text{new}} \leftarrow V_{\text{new}} \cup$ (target node of $e$)
30:       **end if**
31:    **end for**
32:    **return** $V_{\text{new}}$
33: **end procedure**

Table A.5: The **Mining Reward** column indicates the number of minted coins a miner can collect as a reward for mining. The **Underclaimed Blocks** column lists the block heights where the miner(s) claimed a partial amount of the minted coins, along with the corresponding amounts. The **Halving Block** column refers to the block at which the number of minted coins is halved.

| Mining Reward | Underclaimed Blocks | Halving Block |
|---|---|---|
| 50.0 | (124724, 49.98999999), (162839, 49.989752), (162952, 49.91535267), (162973, 49.89494061), (162980, 49.98820406), (162988, 49.99335613), (162992, 49.9505), (163006, 49.96649), (163017, 49.9005), (163063, 49.98649), (163075, 49.9949), (163102, 49.97499935), (163109, 49.98), (163120, 49.81844472), (163228, 49.95109255), (163240, 49.81843189), (163274, 49.98749), (163284, 49.9565), (163368, 49.91699499), (163370, 49.9571021), (163440, 49.9425), (163458, 49.96659523), (163480, 49.949), (163588, 49.9469), (163596, 49.9815), (163611, 49.982), (163615, 49.94788537), (163620, 49.8463), (163623, 49.979), (163626, 49.9957), (163628, 49.97491216), (163633, 49.8215), (163647, 49.8875), (163656, 49.94837683), (163672, 49.9485), (163709, 49.84126551), (163710, 49.86829549), (163714, 49.995), (163719, 49.95048), (163729, 49.988), (163756, 49.98094546), (163791, 49.985), (163830, 49.87694992), (163881, 49.9894), (163889, 49.978), (163903, 49.99599), (163915, 49.9775), (163918, 49.951), (163939, 49.9515), (163975, 49.93509998), (163977, 49.959304), (163985, 49.68647102), (163989, 49.9844), (164003, 49.9683), (164007, 49.868), (164023, 49.973), (164024, 49.992), (164027, 49.9525), (164048, 49.9989), (164049, 49.85630504), (164056, 49.96091352), (164081, 49.9855), (164084, 49.27878274), (164095, 49.9808699), (164098, 49.9678002), (164099, 49.9869998), (164100, 49.95414873), (164105, 49.96910003), (164106, 49.9988), (164127, 49.9915), (164128, 49.9875), (164133, 49.98499846), (164139, 49.9805), (164143, 49.959), (164147, 49.91319995), (164157, 49.96745635), (164159, 49.981), (164162, 49.9865), (164165, 49.969), (164170, 49.9954954), (164173, 49.9885), (164194, 49.9943999), (164196, 49.9955), (164199, 49.9845), (164201, 49.98019), (164203, 49.96155947), (164211, 49.93739999), (164237, 49.97067615), (164246, 48.24219931), (164252, 49.98960918), (164253, 49.9765), (164254, 49.955), (164269, 49.9675), (164280, 49.99296766), (164295, 49.972), (164298, 49.98699997), (164322, 49.83017189), (164334, 49.983), (164365, 49.989), (164380, 49.99204522), (164441, 49.08544057), (164467, 49.9685), (164480, 49.9244), (164528, 49.9725), (164570, 49.88149), (164572, 49.9615), (164615, 49.96691117), (164622, 49.98909039), (164671, 49.996), (164700, 49.94), (164851, 49.84098912), (164921, 49.9209), (164967, 49.9625), (164996, 49.96533227), (165048, 49.9745), (165062, 49.963), (165109, 49.9605), (165124, 49.928), (165177, 49.97458639), (165194, 49.97992), (165283, 49.97383857), (166250, 49.96739476), (166263, 49.895), (166486, 49.9715), (166698, 49.986), (166890, 49.99219564), (167013, 49.938), (167035, 49.9325), (167074, 49.99716399), (167204, 49.9825), (168596, 49.9935), (169464, 49.9785), (169670, 49.9909), (169899, 49.948), (180597, 49.991), (181603, 49.99999), (184450, 49.993), (187478, 49.9945), (187719, 49.997), (188197, 49.994), (189003, 49.99), (189089, 49.99996), (192081, 49.998), (192814, 49.9997), (193020, 49.99949), (193207, 49.9999999), (194078, 49.9985), (195214, 49.99995), (195921, 49.999475), (197935, 49.9975), (198085, 49.99999317), (204342, 49.9965), (204902, 49.9999), (208810, 49.999), (209310, 49.9995) | 210 000 |
| 25.0 | (214251, 24.995), (218683, 24.998), (233700, 24.9988), (236275, 24.9995), (249185, 24.999), (370002, 24.99999999), (404693, 24.99752775), (407187, 24.99823924), (408962, 24.99917564), (410212, 24.99804871) | 420 000 |
| 12.5 | (501726, 0.0), (530371, 12.49994556), (530771, 12.49994622), (531285, 12.49994597), (533906, 12.49999928), (534297, 12.49989145), (534723, 12.49999919), (538597, 12.49999933), (541118, 12.49999931), (541608, 12.49999922), (542549, 12.49999923), (544704, 12.4999993), (546962, 12.4999992), (549077, 12.49999926), (550226, 12.49999929), (553396, 12.49999924), (559931, 12.49999921), (564959, 12.49999925), (619631, 12.49999927), (626205, 12.49999932) | 630 000 |
| 6.25 | | 840 000 |
| 3.125 | | 862 993 |

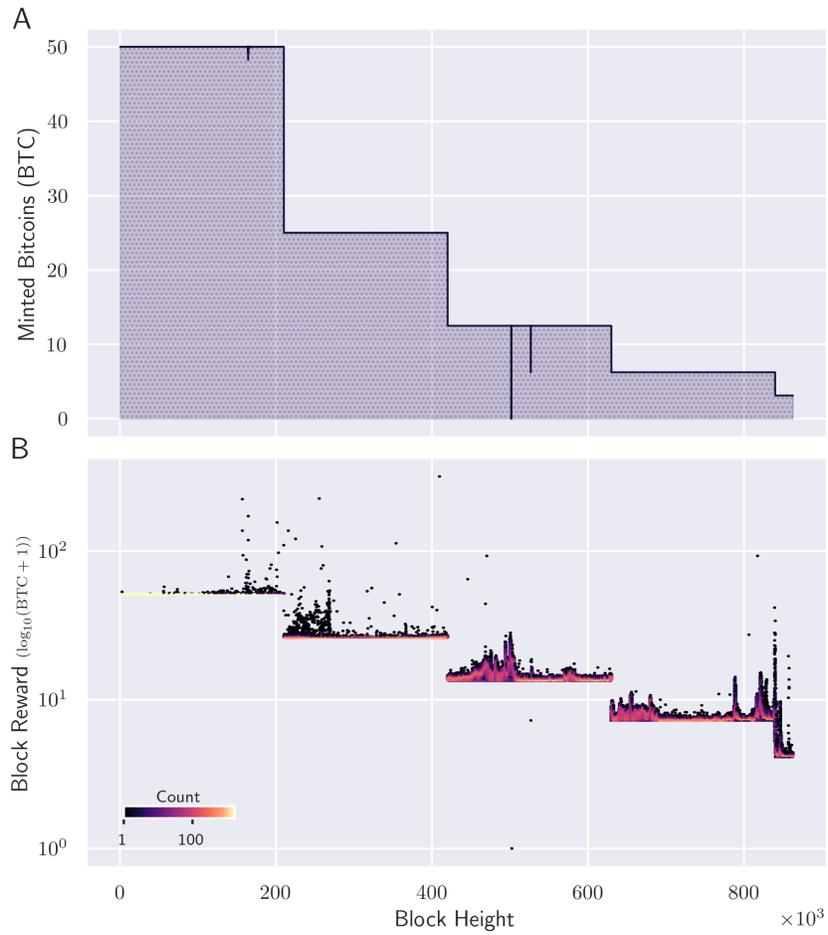

Figure A.6: Bitcoin miner incentives including block rewards and fees. (**A**) With each block, the protocol generates new currency—starting at 50BTC and halving every 210 000 block. To compensate miners, the protocol allows them to claim all the newly minted coins; however, in some blocks (listed in table A.5), miners left a portion—or even all (block 501 726)—of the newly minted coins unclaimed. (**B**) Additionally, miners collect all transaction fees.

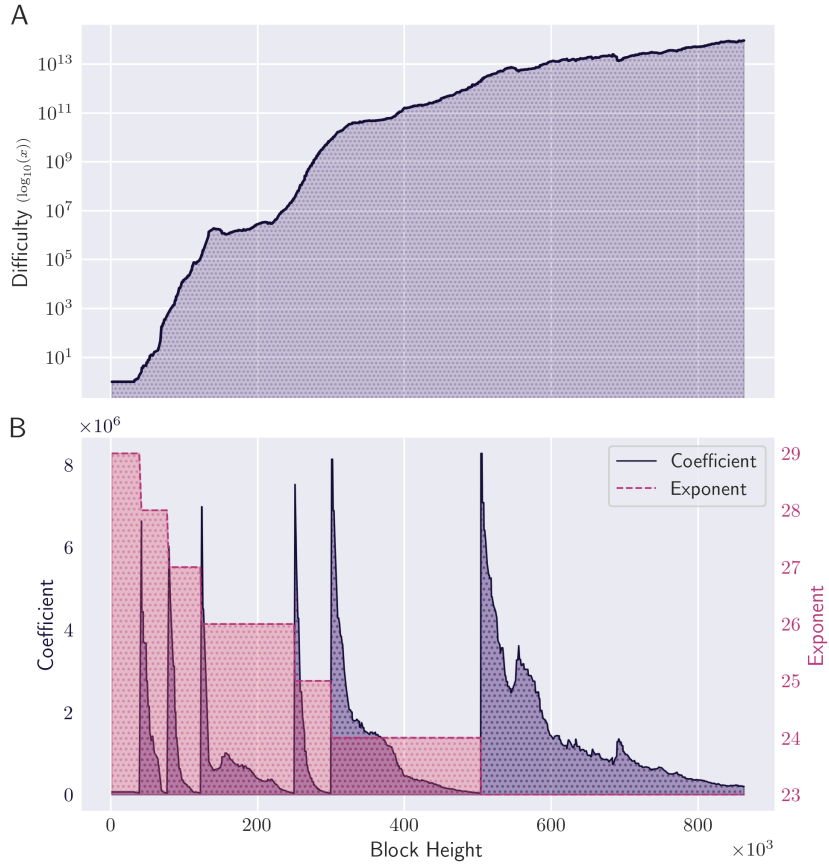

Figure A.7: Bitcoin's decentralized and automatic mining difficulty adjustment mechanism. (**A**) The Bitcoin protocol enforces rules ensuring that blocks are generated at regular intervals by adjusting the mining *difficulty* at approximately every 2 016 block. *Difficulty* is defined as the ratio of the *target* at the Genesis Block to the current *target*, where the *target* is a 256-bit number that sets the upper bound for the hash a miner must compute [13]. (**B**) Each block stores the *target* in a compact form as $coefficient \times 2^{8 \times (exponent - 3)}$.

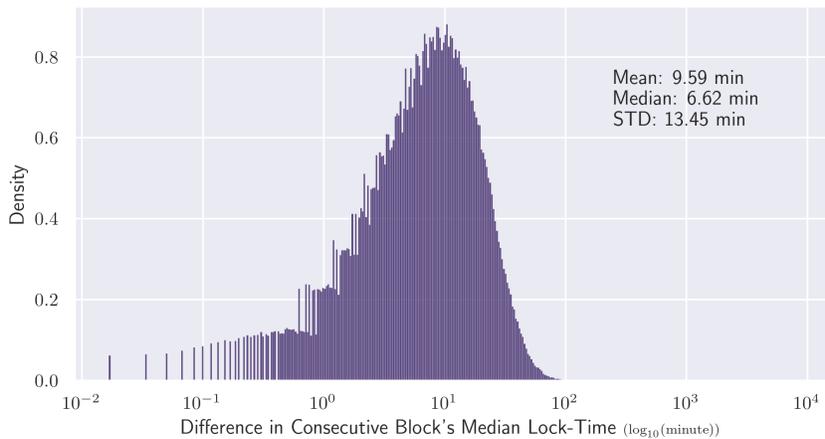

Figure A.8: Distribution of the differences between the lock-times of consecutive blocks. The Bitcoin protocol uses the median of the timestamps of the last 11 blocks as a block's "Median Time" or its lock time [8].

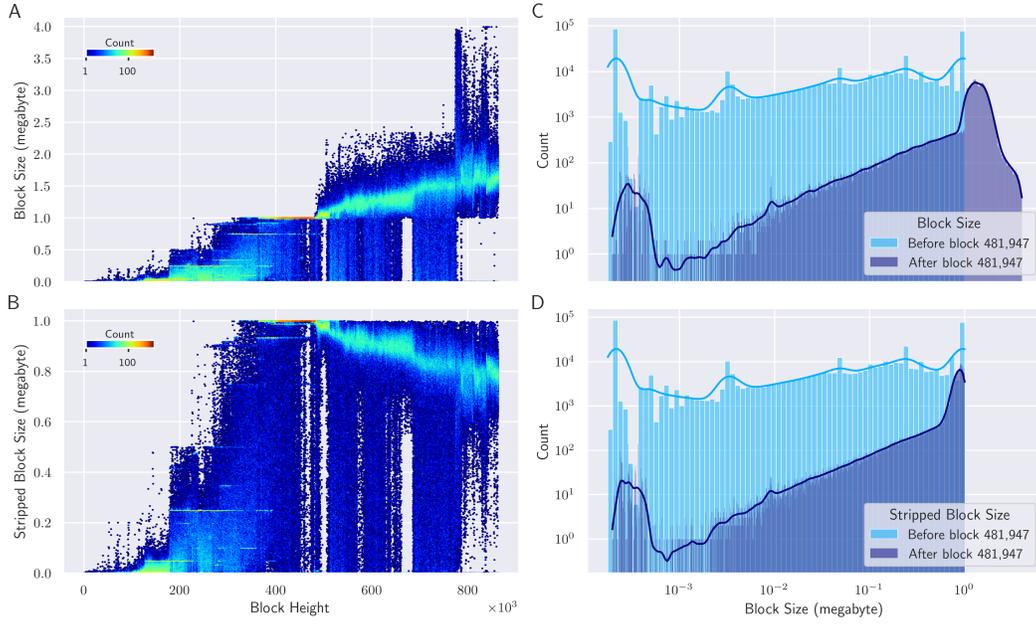

Figure A.9: Bitcoin block size evolution surrounding Segregated Witness (SegWit, BIP 141 [10]), activated at block $481\,824$. The protocol imposes per-block byte limits, hence, each block contains a subset of the transactions from the network's transaction pool at the time of mining, with restrictions on the size of transaction data. SegWit introduced *stripped size* (block size excluding witness data, effectively capped near 1MB) and a block weight limit $\leq 4\,000\,000$ Weight Units, enabling total block sizes $>$1MB (first observed: block $481\,947$). The first block where size and stripped size differ is block $481\,824$, where size is $989\,323$ bytes and stripped size is $988\,519$ bytes. Plots compare longitudinal block size (**A**) and stripped block size (**B**), highlighting the increase in block size after block $481\,947$ and the capped nature of stripped size. Panels **C** and **D** compare the distributions of block size and stripped size, respectively, before and after block $481\,947$. Stripped size for blocks prior to SegWit is reported as equal to block size by the Bitcoin Core application; hence, we include it here for completeness. table A.6 summarizes the key statistical moments.

Table A.6: Block size summary statistics; see Fig. A.9.

|  | Before $481\,947$ | | | | | After $481\,947$ | | | | |
| --- | --- | --- | --- | --- | --- | --- | --- | --- | --- | --- |
|  | Min | Max | Avg | Std | Sum | Min | Max | Avg | Std | Sum |
| Size (MB) | 0.0 | 1.0 | 0.2 | 0.3 | $130\,778$ | 0.0 | 3.9 | 1.2 | 0.4 | $472\,860$ |
| Stripped Size (MB) | 0.0 | 1.0 | 0.2 | 0.3 | $130\,777$ | 0.0 | 0.9 | 0.7 | 0.2 | $287\,079$ |

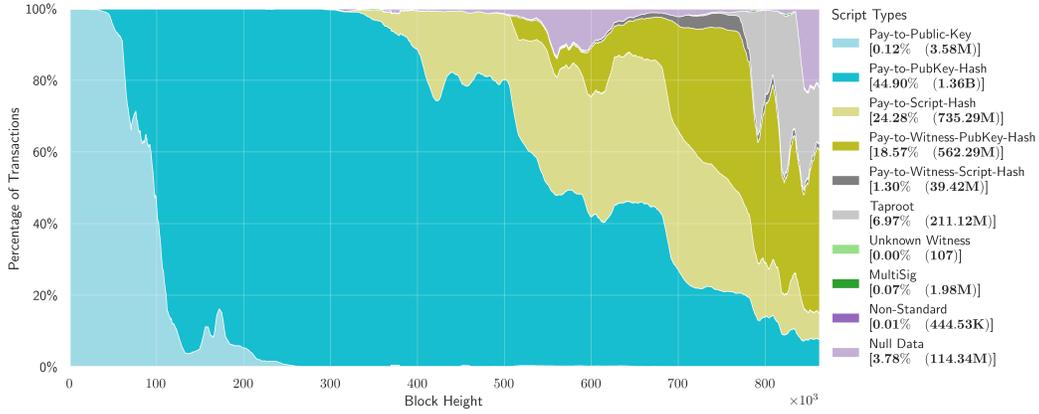

Figure A.10: Longitudinal distribution of script type usage, illustrating the percentage of scripts per block for common types and null data (see Fig. 3(G-H) for further detail). The plot shows an early dominance of Pay-to-Public-Key (P2PK), later shifting to Pay-to-Public-Key-Hash (P2PKH), and more recently, an increasing adoption of Pay-to-Witness-Public-Key-Hash (P2WPKH). "Null Data" refers to scripts primarily used to encode arbitrary data on the blockchain, typically by embedding a small amount of data in exchange for a transaction fee.

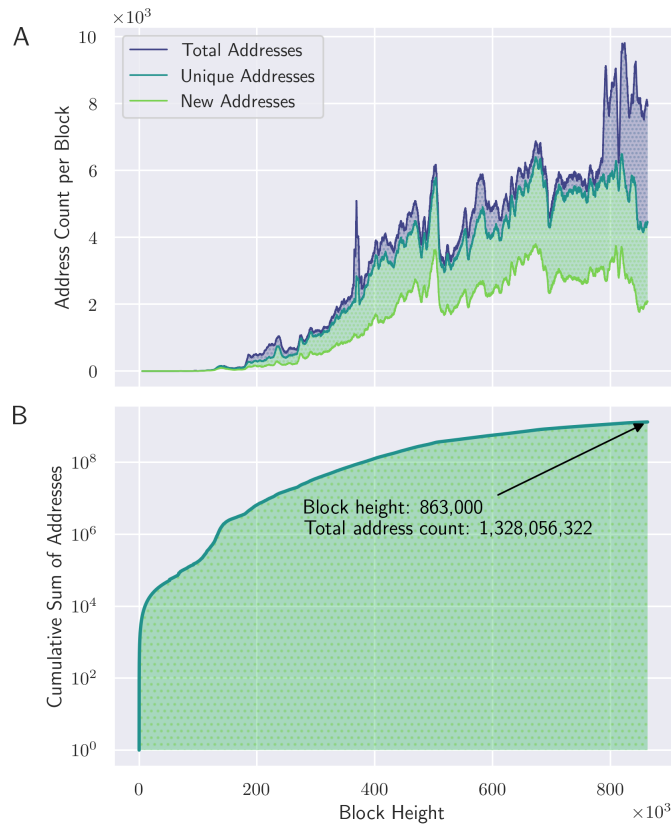

Figure A.11: Panel **A** plots per-block address counts, differentiating total addresses referenced, unique addresses, and new addresses (observed for the first time in that block). Panel **B** plots the cumulative sum of new addresses over time. Address counts are smoothed using a 5 000-block moving average.

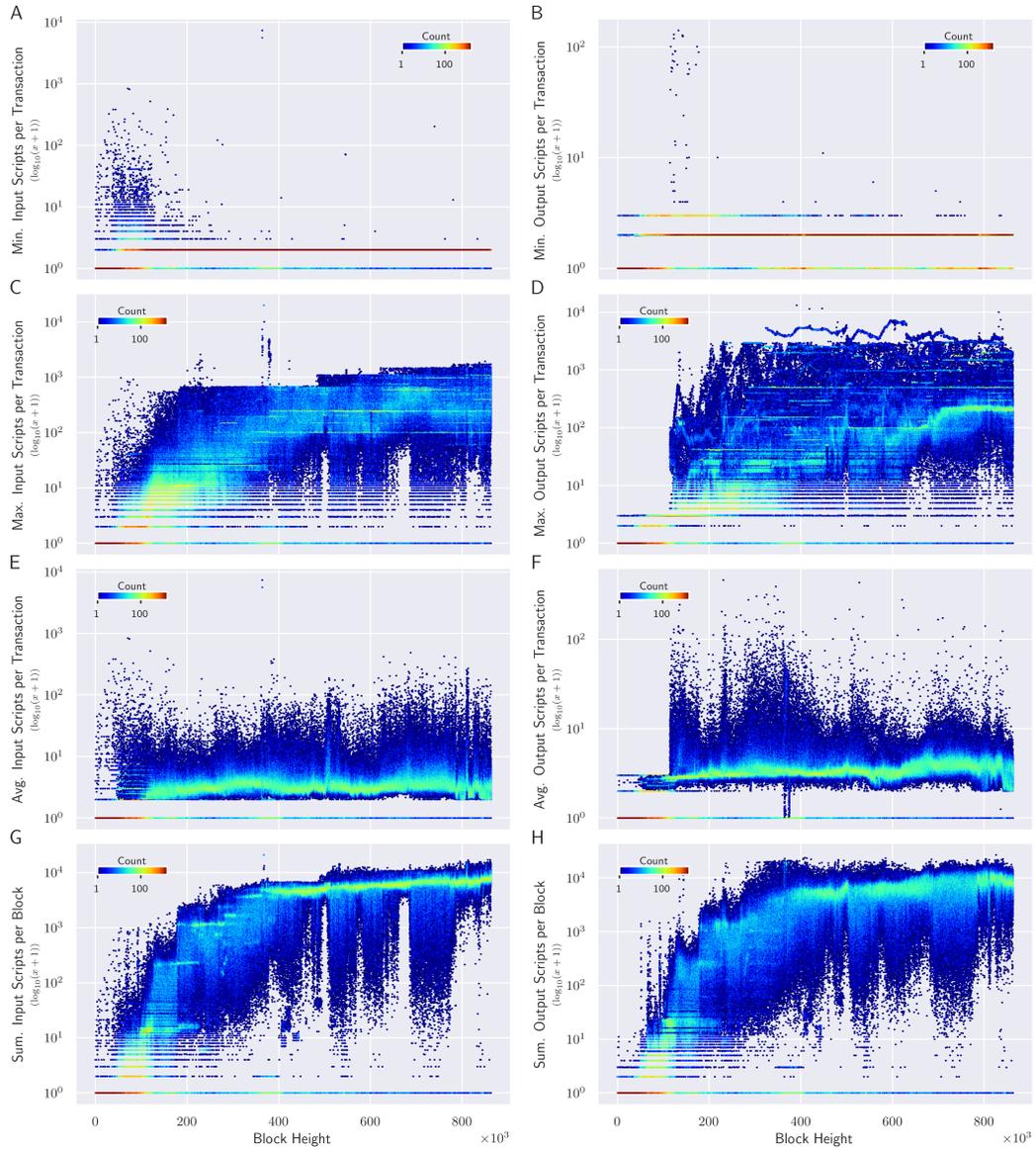

Figure A.12: Per-block statistics on the count of TxIn and TxOut per Tx. Panels (**A**, **C**, **E**, **G**) detail, for each block, the minimum, maximum, and average number of TxIn per transaction, alongside the total number of TxIn in that block. Similarly, panels (**B**, **D**, **F**, **H**) show corresponding statistics for the number of TxOut per Tx and the total number of TxOut in that block. See Fig. A.13 for related BTC value statistics per block.

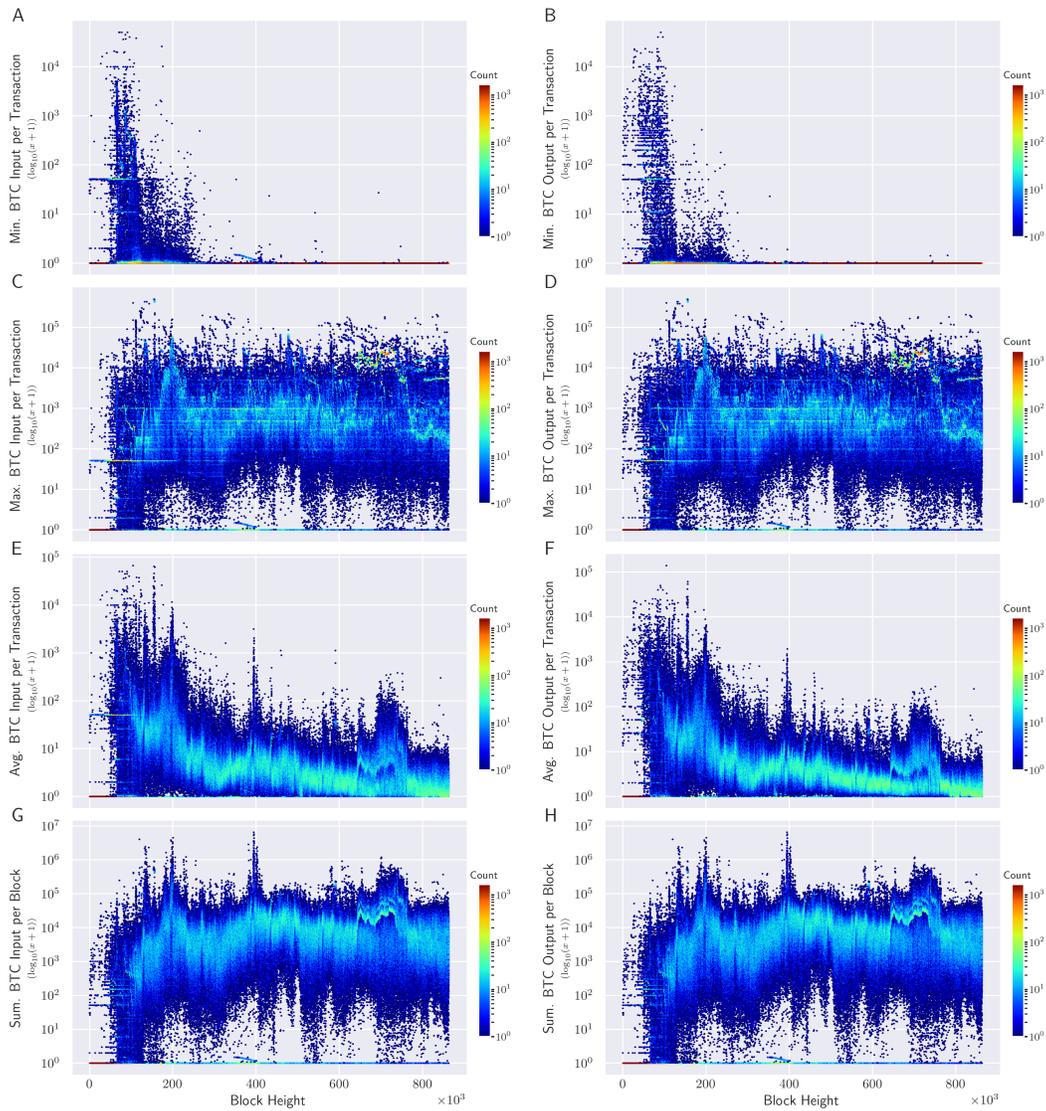

Figure A.13: Per-block BTC value statistics for transaction TxIn and TxOut. Panels (**A**, **C**, **E**, **G**) detail the minimum, maximum, average, and sum of BTC values for TxIn within each block, while panels (**B**, **D**, **F**, **H**) show the corresponding statistics for TxOut. The corresponding distributions are similar because, under Bitcoin's protocol, any TxIn value not explicitly directed to recipient TxOut or designated as transaction fees becomes permanently unspendable (Fig. A.14). The TxOut values plotted here represent only amounts paid to recipients, excluding transaction fees (see also Fig. A.14).

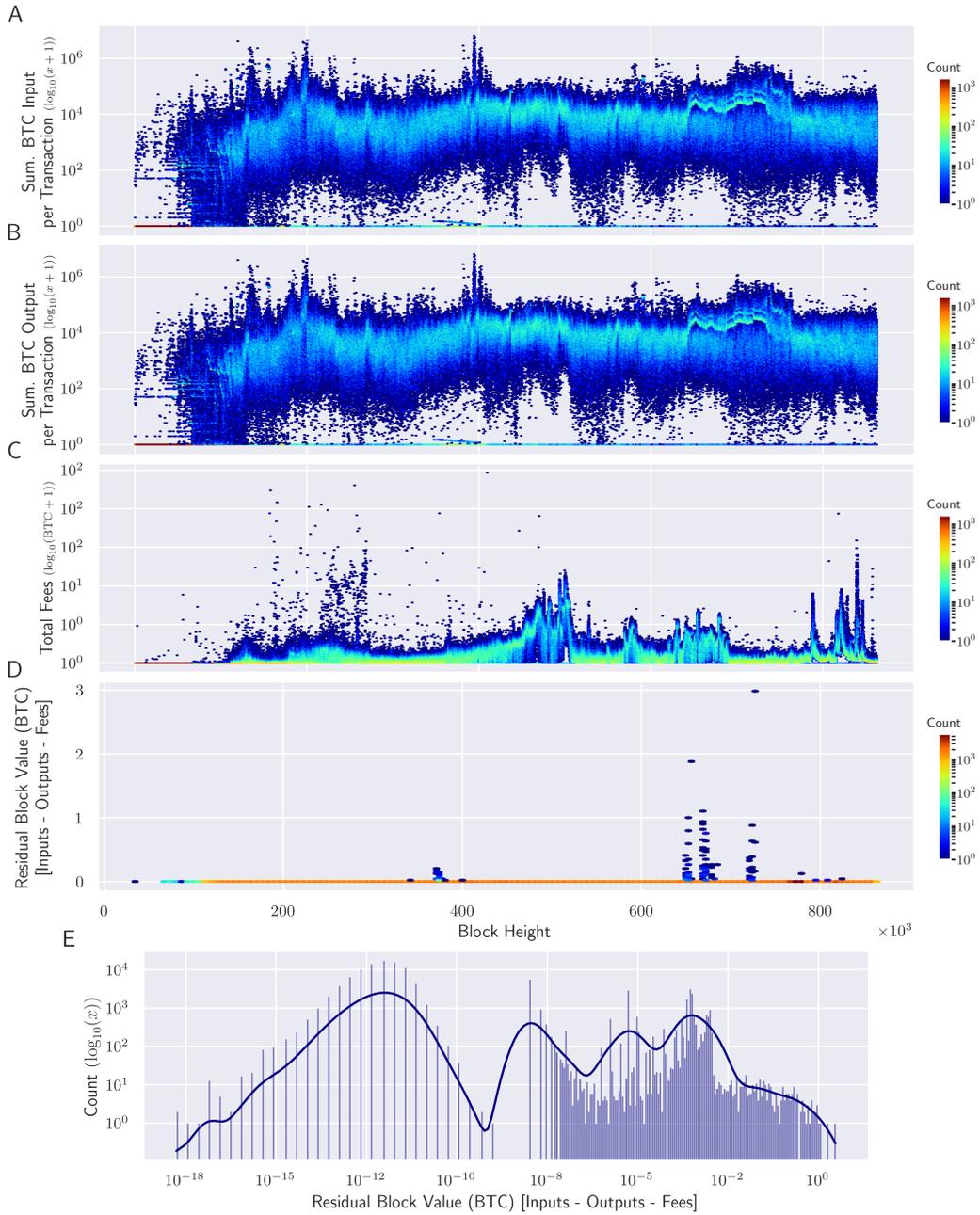

Figure A.14: Theoretically, the total value of spent inputs (unlocked previous outputs, $\Sigma_{\text{in}}$) in a transaction should equal the sum of its output values ($\Sigma_{\text{out}}$) plus the transaction fee paid to the miner ($f$). However, in 116,067 blocks, $\Sigma_{\text{in}} - \Sigma_{\text{out}} - f > 0$, non-zero residual, summing to 40.554 05 BTC across these instances. This residual represents value not accounted for by outputs or fees, effectively lost within these transactions. Panels **A**, **B**, and **C** respectively plot the total value of $\Sigma_{\text{in}}$, $\Sigma_{\text{out}}$, $f$ over the blockchain. Panel **D** highlights the blocks containing transactions with non-zero residuals, and Panel **E** plots the distribution of these residual values. The counts for different ranges of residual value $x$ are: 92,536 instances where $0 < x \leq 1 \times 10^{-8}$ BTC, 23,528 instances where $1 \times 10^{-8} < x \leq 1$ BTC, and 3 instances where $1 < x$ BTC.

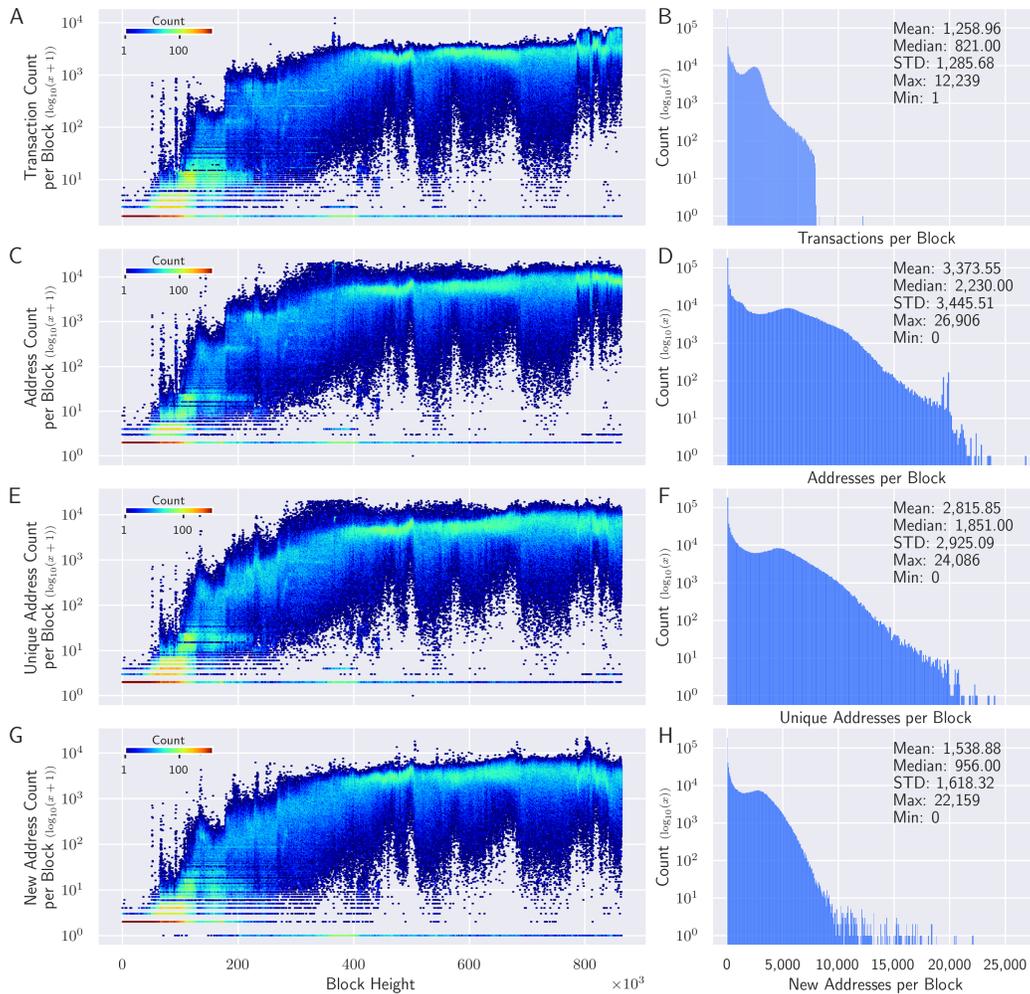

Figure A.15: Per-block Tx and address count statistics. Panels (**A**, **C**, **E**, **G**) longitudinally plot per-block counts, while panels (**B**, **D**, **F**, **H**) present their corresponding distributions, annotated with minimum, maximum, mean, median, and standard deviation. Blocks with only one transaction are "empty blocks", containing only the coinbase Tx (generated when a miner succeeds in mining a block before including transactions from the *mempool*). Notably, block 501 726 is the only one with zero addresses, as its entire reward was unclaimed (i.e., TxOut value was 0) and it contained only a single non-standard script output. The plots illustrate the expected significant correlation between Tx count and address counts per block: Tx count vs. total addresses per block (Pearson correlation $\rho = 0.8739$, panels **A** and **C**), Tx count vs. unique addresses per block ($\rho = 0.7705$, panels **A** and **E**), Tx count vs. new addresses per block ($\rho = 0.7942$, panels **A** and **G**), total addresses per block vs. unique addresses per block ($\rho = 0.9524$, panels **C** and **E**), and total addresses per block vs. new addresses per block ($\rho = 0.8896$, panels **C** and **G**). See Fig. A.11 for cumulative and rolling mean plots of address counts per block.

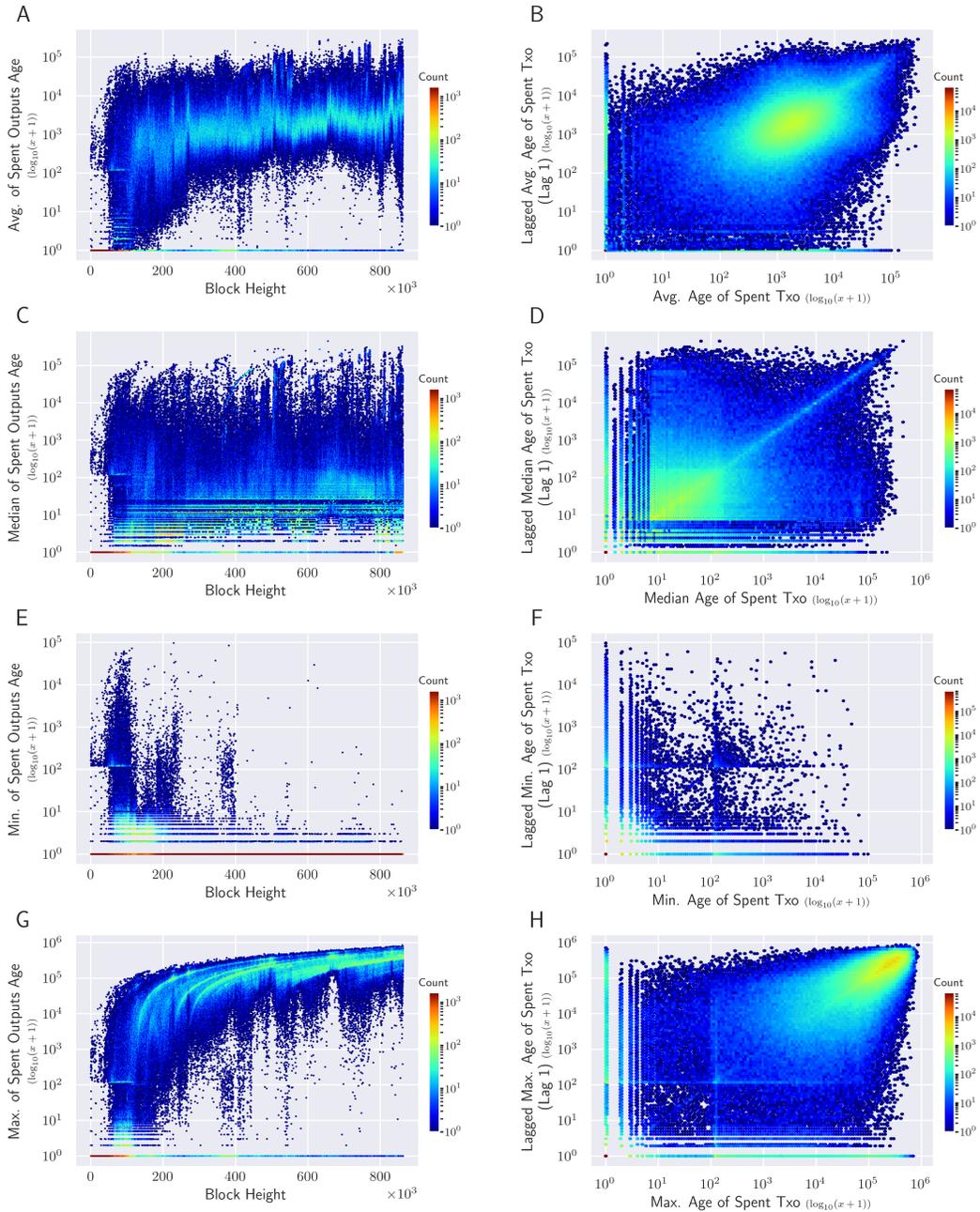

Figure A.16: Age (coin dormancy) of spent transaction outputs (TxOut). Panels (**A**, **C**, **E**, **G**) longitudinally plot the per-block average, median, minimum, and maximum age for TxOut spent in that block (i.e., used as TxIn). Panels (B, D, F, H) illustrate the relationship between the age of spent TxOut and their corresponding lag-1 age values. Age is defined as the difference in block height between TxOut creation and spending.

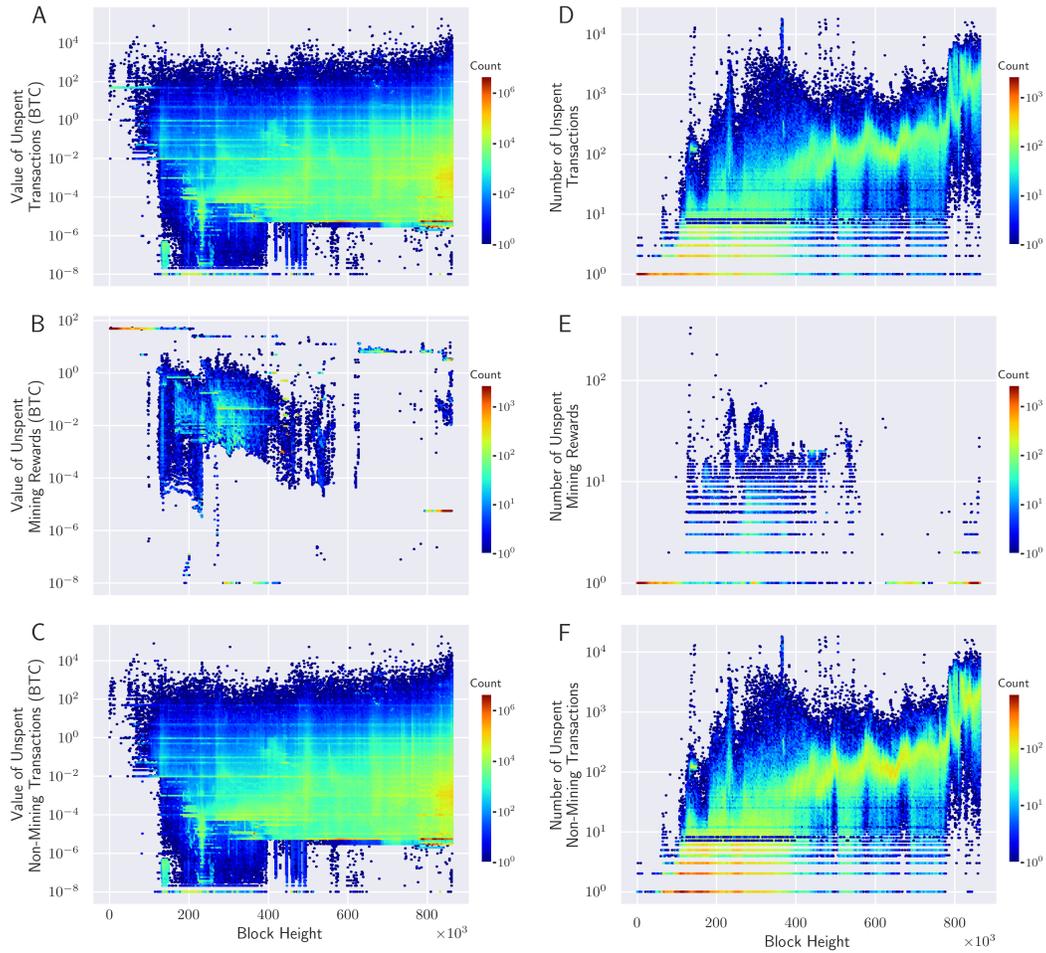

Figure A.17: Plots illustrate longitudinal details of UTxOs up to block 863 000, indicating the block height at which each UTxO was created. Panels **A**-**C** plot BTC values, while panels **D**-**F** plot their corresponding counts. Panel **A** shows the longitudinal BTC value of all UTxOs created per block, including both coinbase and non-coinbase TxOut. Panels **B** and **C** provide a breakdown of panel **A**: panel **B** plots the value of coinbase UTxOs created per block, and panel **C** plots the value of non-coinbase UTxOs created per block. Correspondingly, panels **D**, **E**, and **F** plot the longitudinal counts of these UTxOs created per block and remaining unspent: panel **D** for all UTxOs, panel **E** for coinbase UTxOs, and panel **F** for non-coinbase UTxOs.

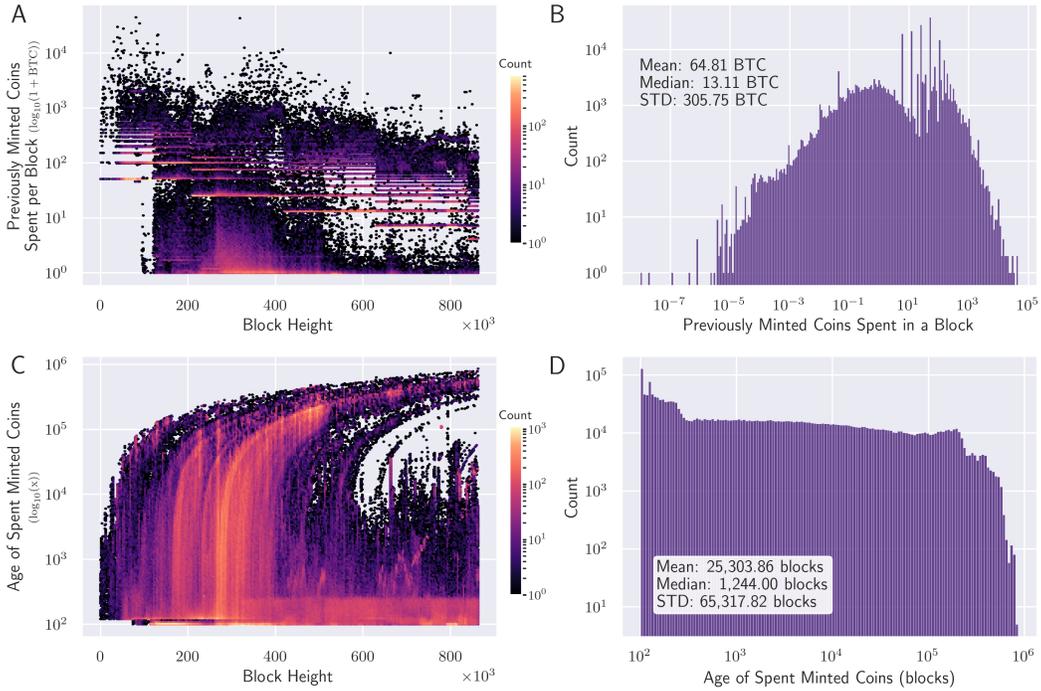

Figure A.18: Spending patterns of previously minted coins. Panel **A** provides a longitudinal plot of the total BTC amount from minted coins spent per block, excluding any fees that might have been part of prior coinbase Tx TxOut. Panel **B** shows the distribution of these per-block spent BTC values. Panel **C** provides a longitudinal plot of the age of these spent minted coins at the time of spending, defined as the difference in block height between their creation (minting) and expenditure. Panel **D** presents the distribution of these ages.

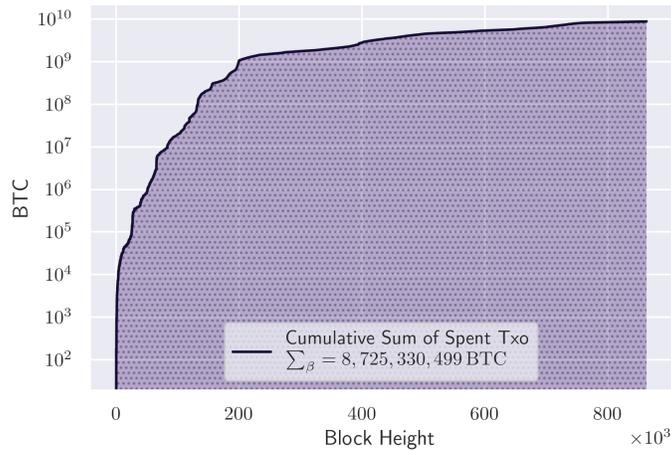

Figure A.19: Cumulative sum of traded BTC, defined in terms of the spent transaction outputs.

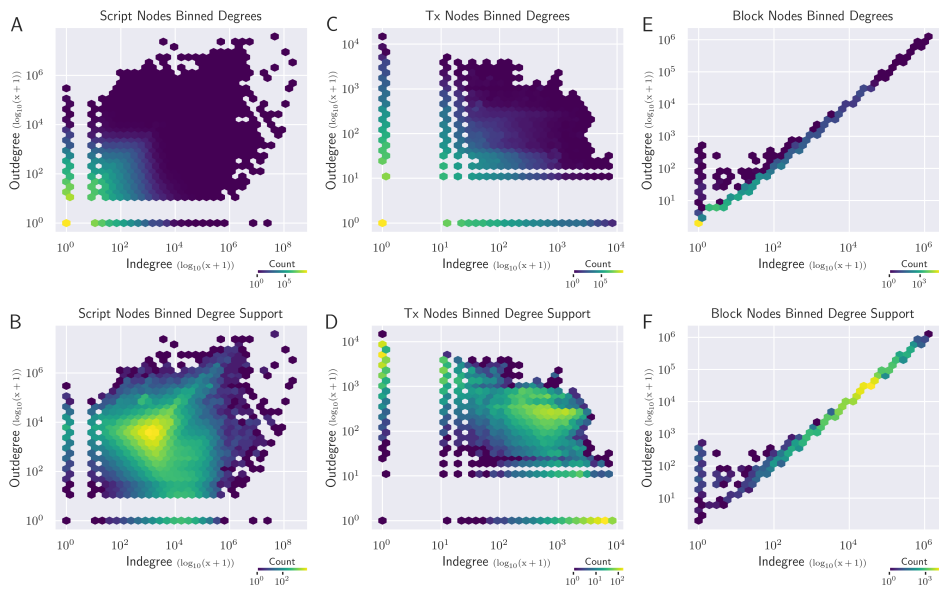

Figure A.20: The figure shows the binned in and out degrees of nodes, with bin size of 10. panels A, C, and E plot the count of degree pairs in each bin, and panels B, D, and F plot support of each bin where if there is at least one value in a bin its count will be 1 otherwise it will be zero. The degree counts include all types of in and out relationships of a node. Of 860

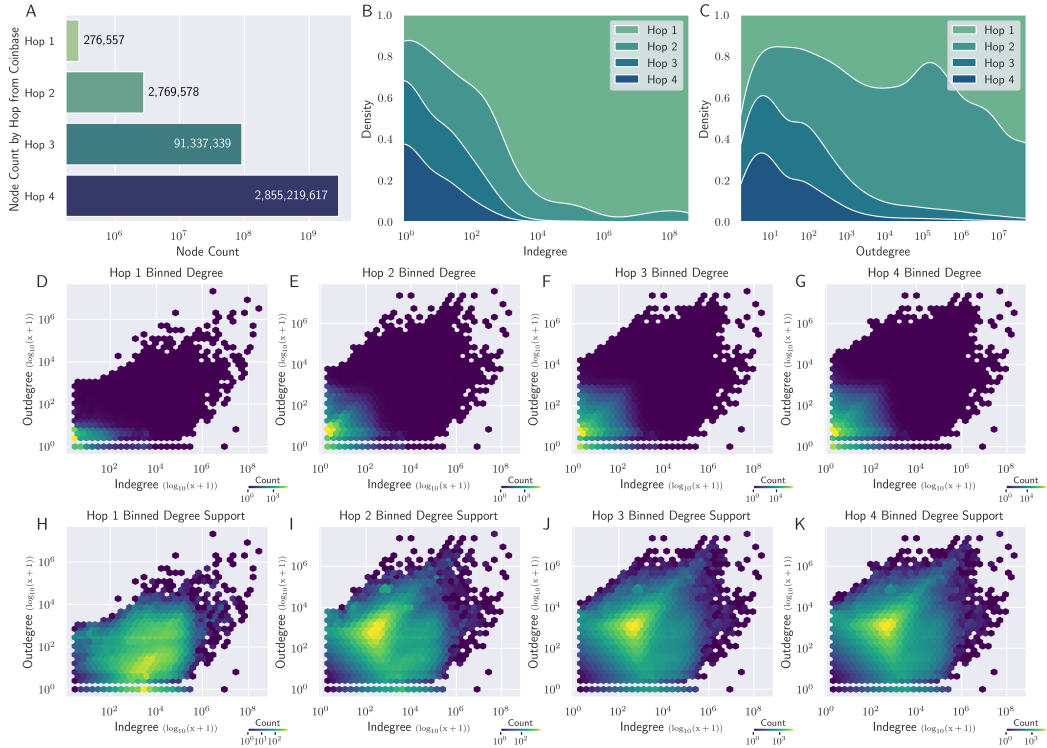

Figure A.21: Degree characteristics within the $k$-hop neighborhood ($k = 1$ to $4$) of the *Coinbase* node, considering only *script* nodes and *script-to-script* edges. Degrees reflect script-to-script edges originating from nodes at the specified hop distance, and no restriction on target node's distance from the *Coinbase* node. Node degrees are binned using a window size of $10$. (**A**) Number of unique script nodes at each hop distance; *Hop 1* nodes ($276\,557$) typically represent script addresses receiving newly minted coins (e.g., miner or mining pool scripts), while *Hop 2* nodes receive funds from *Hop 1* scripts (e.g., mining pools not paying miners from the coinbase Tx). (**B**, **C**) Overall frequency distributions of binned indegrees (**B**) and outdegrees (**C**) for all unique script nodes within 4 hops. (**D-G**) Joint degree density plots (binned indegree vs. outdegree) for script nodes at hops 1-4 respectively; color intensity corresponds to node count per hexbin. (**H-K**) Joint degree support plots for script nodes at hops 1-4 respectively; a colored bin indicates the existence ($\geq 1$ node) of that degree combination. While the density plots (**D-G**) show the vast majority of nodes have low degrees, the support plots (**H-K**) reveal a much wider range, as the majority of nodes have significantly higher degree ($10 \times 10^2$-$10 \times 10^4$), suggesting a significant heterogeneity in the degree distribution. Note that these plots represent a snapshot at a fixed block height; longitudinal comparison could provide in-depth insights about the evolving trends and predictive capacity.

Table A.7: Entropy of block, Tx, and script nodes, where normalized Shannon entropy is defined as $H_n := \frac{H}{H_{max}}$, with maximum entropy $H_{max} := \log(M)$, $H := -\sum(p * \ln(p))$, and density is $D := \frac{E}{N \times (N-1)}$. Block nodes connect Txs and scripts, hence, despite being fewer nodes, they are densely connected. Contrary, since Tx nodes represent unique transactions that are connected to the script nodes they define and to the blocks where their inputs and outputs were created and used, respectively, hence they are least densely connected. Script nodes, despite being numerous, form a sparse network. However, the low normalized entropy for both the Tx and script nodes indicates less diversity in the number of edges these nodes have in the network. In other words, the low entropy values indicates a more uniform and predictable topology or a low level of disorder or uncertainty (in terms of the number of connections) for most Tx and script nodes.

|  | Statistic | Block Nodes | Tx Nodes | Script Nodes |
|---|---|---|---|---|
|  | Node Count ($N$) | 863 000 | 1 082 275 341 | 1 311 600 327 |
|  | Edge Count ($E$) | 13 238 562 732 | 3 582.643 395 | 9 593 807 363 |
|  | Density ($D$) | 0.017 775 42 | $3.058 \times 10^{-9}$ | $5.576 \times 10^{-9}$ |
| Indegree | Mean | 15 340.165 4 | 3.310 2 | 7.314 |
|  | Standard deviation | 20 793.345 7 | 52.433 | 12 152.895 |
|  | Distinct Values ($M$) ($H$) | 54 399 | 7 641 | 33 498 |
|  | Raw Entropy ($H$) | 9.512 978 | 0.998 759 | 1.490 987 |
|  | Max Entropy ($H_{max}$) | 10.904 101 | 8.941 284 | 10.419 241 |
|  | Normalized Entropy ($H_n$) | 0.872 422 | 0.111 702 | 0.143 099 |
| Outdegree | Mean | 15 344.789 6 | 6.617 3 | 14.621 7 |
|  | Standard deviation | 20 802.611 3 | 23.582 | 2 530.272 8 |
|  | Distinct Values ($M$) ($H$) | 54 382 | 2 908 | 22 371 |
|  | Raw Entropy ($H$) | 9.515 625 | 1.240 856 | 1.784 836 |
|  | Max Entropy ($H_{max}$) | 10.903 788 | 7.975 221 | 10.015 521 |
|  | Normalized Entropy ($H_n$) | 0.872 690 | 0.155 589 | 0.178 207 |
| Total | Mean | 30 684.955 | 9.927 6 | 21.936 2 |
|  | Standard deviation | 41 595.954 | 57.438 2 | 12 770.357 3 |
|  | Distinct Values ($M$) ($H$) | 77.819 | 7 833 | 48 065 |
|  | Raw Entropy ($H$) | 9.664 467 | 1.949 118 | 2.591 089 |
|  | Max Entropy ($H_{max}$) | 11.262 141 | 8.966 101 | 10.780 310 |
|  | Normalized Entropy ($H_n$) | 0.858 138 | 0.217 387 | 0.240 354 |